\documentclass[10pt,twocolumn,letterpaper]{article}

\usepackage{makecell}
\usepackage[pagenumbers]{wacv} 
\usepackage[accsupp]{axessibility}

\usepackage{multirow}
\usepackage{capt-of}
\usepackage{cuted}
\newcommand{\rev}[1]{#1}

\usepackage{pgfplots}
\usepackage{placeins}
\pgfplotsset{compat=1.18}

\DeclareCaptionFont{revisionblue}{\color{black}}

\providecommand{\nolinenumbers}{}
\providecommand{\linenumbers}{}

\definecolor{wacvblue}{rgb}{0.21,0.49,0.74}
\usepackage[pagebackref,breaklinks,colorlinks,allcolors=wacvblue]{hyperref}

\def\wacvPaperID{874} 
\def\confName{WACV}
\def\confYear{2027}

\title{Spatiotemporal Dynamics-Aware Industrial Anomaly Detection and Reasoning}

\author{
S-Wei Chen\textsuperscript{*} \quad
Siangyu Yang\textsuperscript{*} \quad
Yi-Chen Jan\textsuperscript{*} \quad
Chun-Yi Lee \quad Shao-Yuan Lo \\
National Taiwan University \\
{\tt\small
\{b11902027, b12902055, b11902057, cylee, sylo\}@csie.ntu.edu.tw
}
}

\begin{document}
\maketitle
\begingroup
\renewcommand{\thefootnote}{*}
\footnotetext{Equal contribution.}
\endgroup
\begin{abstract}
Industrial Anomaly Detection (IAD) is a long-established problem that aims to detect defects in industrial products during manufacturing for quality control. Most existing studies focus on static visual patterns as the basis for detection. However, in real manufacturing environments, many anomalies are dynamics-based, such as irregular movements that cannot be perceived from static cues alone. We observe that even recent Multimodal Large Language Model (MLLM)-based approaches fall short in identifying such dynamic anomalies due to limited temporal modeling capability. To address this, we introduce MUSTARD: MUltimodal SpatioTemporal Anomaly Reasoning and Detection, a framework that equips MLLMs with fine-grained spatiotemporal dynamics awareness for detecting and reasoning about dynamics-based defects.
MUSTARD introduces optical-flow-guided keyframe selection and augmented modalities to capture critical motion irregularities through local motion patterns and global dynamic evolution. Its normal pattern-based reasoning pipeline requires only a few normal reference samples as the detection basis, avoiding any model updates. This enables efficient adaptation across various products and provides high decision interpretability.
MUSTARD is one of the first MLLM-based frameworks for dynamics-based IAD. Experiments demonstrate its superior performance in detecting spatiotemporal anomalies and providing plausible explanations. Our code is available at \url{https://github.com/racercode/Mustard}
\vspace{-0.2cm}
\end{abstract}
     
\section{Introduction}
\label{sec:intro}

\begin{figure}
    \centering
    \includegraphics[width=0.95\linewidth]{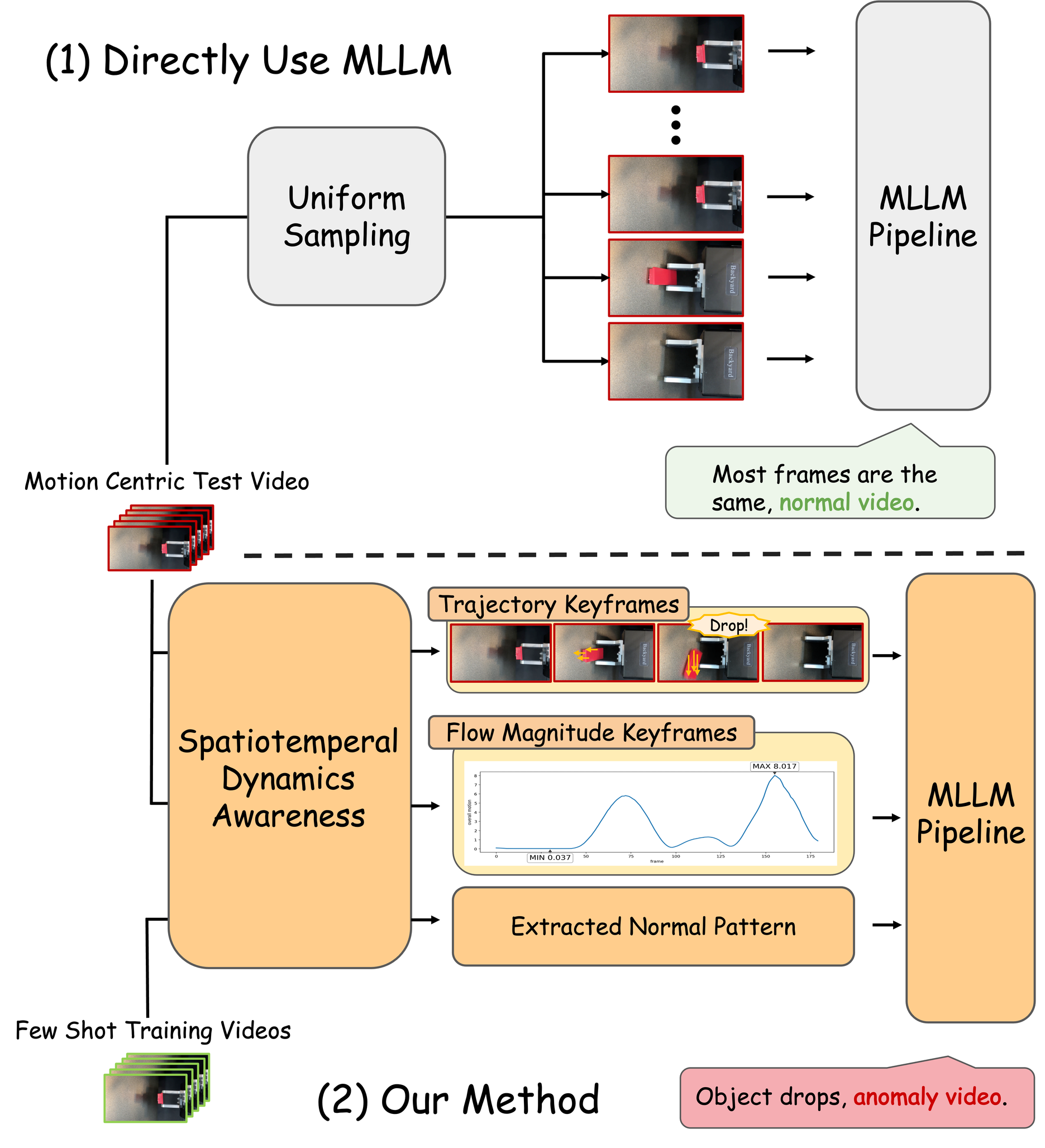}
    \caption{
    Comparison between the direct use of an MLLM and the proposed method on an anomalous test video (a degaussing magnet). 
    (1) Directly Use MLLM: fails to recognize that the magnet falls off the metal surface. 
    (2) Our Method: leverages spatiotemporal information to correctly interpret the magnet's drop from the surface.
    }
    \label{fig:teaser}
    \vspace{-0.6cm}
\end{figure}

Industrial Anomaly Detection (IAD) is a crucial problem that aims to automatically inspect product defects in manufacturing environments. Most existing IAD benchmarks~\cite{bergmann2019mvtec, heckler2025mvtec, zou2022spot, bergmann2022beyond, wang2024real}, and approaches~\cite{zavrtanik2021draem,you2022unified,cao2024adaclip,jeong2023winclip} focus on static visual defects, e.g., a missing chunk on a candle or a bulge on the side of a bowl. These static anomalies can be identified from a single image. However, in real production lines, many anomalies are driven by dynamic irregularities, such as irregular rotations, abnormal vibrations, and unexpected component interactions, which cannot be captured from static cues alone. Inspecting such dynamics-based anomalies requires physical operations. For example, a fan may appear visually flawless from a static observation, but we need to turn it on to see whether it rotates smoothly or gets stuck. Although modern smart factories have been able to leverage robot arms to automate these tests, we still lack an effective IAD algorithm capable of inspecting dynamic anomalies from the resulting test videos, which requires fine-grained spatiotemporal dynamic understanding. Advancing such robot perception capabilities is crucial for seamless integration with robotic arm operations, enabling a fully automated smart manufacturing process.

Existing IAD approaches face fundamental limitations. Traditional convolutional network–based methods employ self-supervised pretext tasks, such as image reconstruction, to model normal patterns ~\cite{zavrtanik2021draem,you2022unified,lo2022adversarially}. CLIP-based frameworks ~\cite{jeong2023winclip,cao2024adaclip,zhou2024anomalyclip} leverage vision-language semantic alignment to achieve few-shot or zero-shot IAD. Nevertheless, both streams consider only static defects and thus rely primarily on static visual attributes. In addition, they provide only anomaly scores, without interpretable rationales, to users. While recent advances in Multimodal Large Language Models (MLLMs) have shown remarkable capabilities across various vision tasks~\cite{liu2023visual,bai2025qwen2}, they still fall short in dynamics-based IAD~\cite{yuan2026vad}. We observe that a key reason stems from dynamic patterns unfolding over multiple video frames. For instance, determining whether a gear rotates correctly requires capturing motion across consecutive frames; a single snapshot cannot reveal whether the gear is stuck, rotating incorrectly, or moving at an improper speed. Similarly, detecting abnormal vibrations or timing misalignments necessitates capturing the temporal evolution of system behavior. However, current MLLMs tend to overlook or fail to prioritize critical anomalous dynamics. We reveal that even video-oriented MLLMs~\cite{lin-etal-2024-video,zhang-etal-2023-video} lack an effective mechanism to explicitly model frame-to-frame temporal changes (see Fig.~\ref{fig:teaser}-(1)), making it difficult to detect subtle variations such as gradual slowdowns or slight trajectory deviations. Moreover, most MLLMs employ uniform frame sampling to reduce computational cost. Although uniform sampling may be sufficient for general video understanding tasks, it fundamentally limits their ability to capture fine-grained temporal anomalies.

To address this challenge, we posit that optical flow, which encodes pixel-level motion between consecutive frames, provides a rich yet underutilized signal for dynamics-based IAD. Flow-based directional arrows reveal local movement trajectories, enabling the tracking of object positions and rotations over time. The flow magnitude represents global motion speed, and its variation highlights critical state transitions, such as when components start moving, stop, or change velocity. By converting optical flow into interpretable visual and statistical representations, we provide MLLMs with explicit temporal dynamics information that complements static frame content.

In this paper, we propose MUSTARD: MUltimodal SpatioTemporal Anomaly Reasoning and Detection, a novel framework that equips MLLMs with optical flow-guided spatiotemporal dynamics awareness, enabling accurate detection and reasoning of dynamic anomalies (see Fig.~\ref{fig:teaser}-(2)). MUSTARD features three key innovations: (1) an optical flow-guided keyframe selection strategy that precisely samples frames containing critical motion irregularities; (2) two auxiliary optical flow–augmented modalities, including trajectory-overlaid video frames annotated with local motion patterns using directional arrows and a flow magnitude curve representing global dynamic evolution; and (3) a normal pattern–based reasoning pipeline that extracts normal patterns from a small set of reference data as the decision basis, providing plausible anomaly explanations. The entire pipeline adheres to the one-class IAD (a.k.a. semi-supervised IAD) setting, in which only normal training samples are available, and no anomaly knowledge is leaked.

In addition, MUSTARD is a few-normal-shot~\cite{yang2024follow} method that uses only a small number of normal samples from the training set as references to extract normal patterns for IAD. This eliminates the need for expensive full-shot training on the entire dataset and does not require any model weight updates, enabling efficient adaptation to various industrial product lines in real manufacturing scenarios. We evaluate our method on Phys-AD~\cite{li2025towards} and IPAD~\cite{ipad}, the two pioneering dynamics-based IAD benchmarks available in the literature. Extensive experiments demonstrate MUSTARD’s state-of-the-art detection performance and reasoning ability. The main contributions are threefold:
\begin{itemize}
    \item We propose one of the first MLLM-based frameworks for dynamics-based IAD, addressing challenges beyond static IAD and augmenting automation in manufacturing.
    \item We introduce a novel optical-flow-guided spatiotemporal dynamics awareness method, enabling accurate detection and reasoning of dynamic anomalies.
    \item The proposed normal pattern-based reasoning pipeline allows efficient adaptation to various industrial products, which is highly practical for real production lines.
\end{itemize}

\begin{figure*}
    \centering
    \includegraphics[width=0.85 \linewidth]{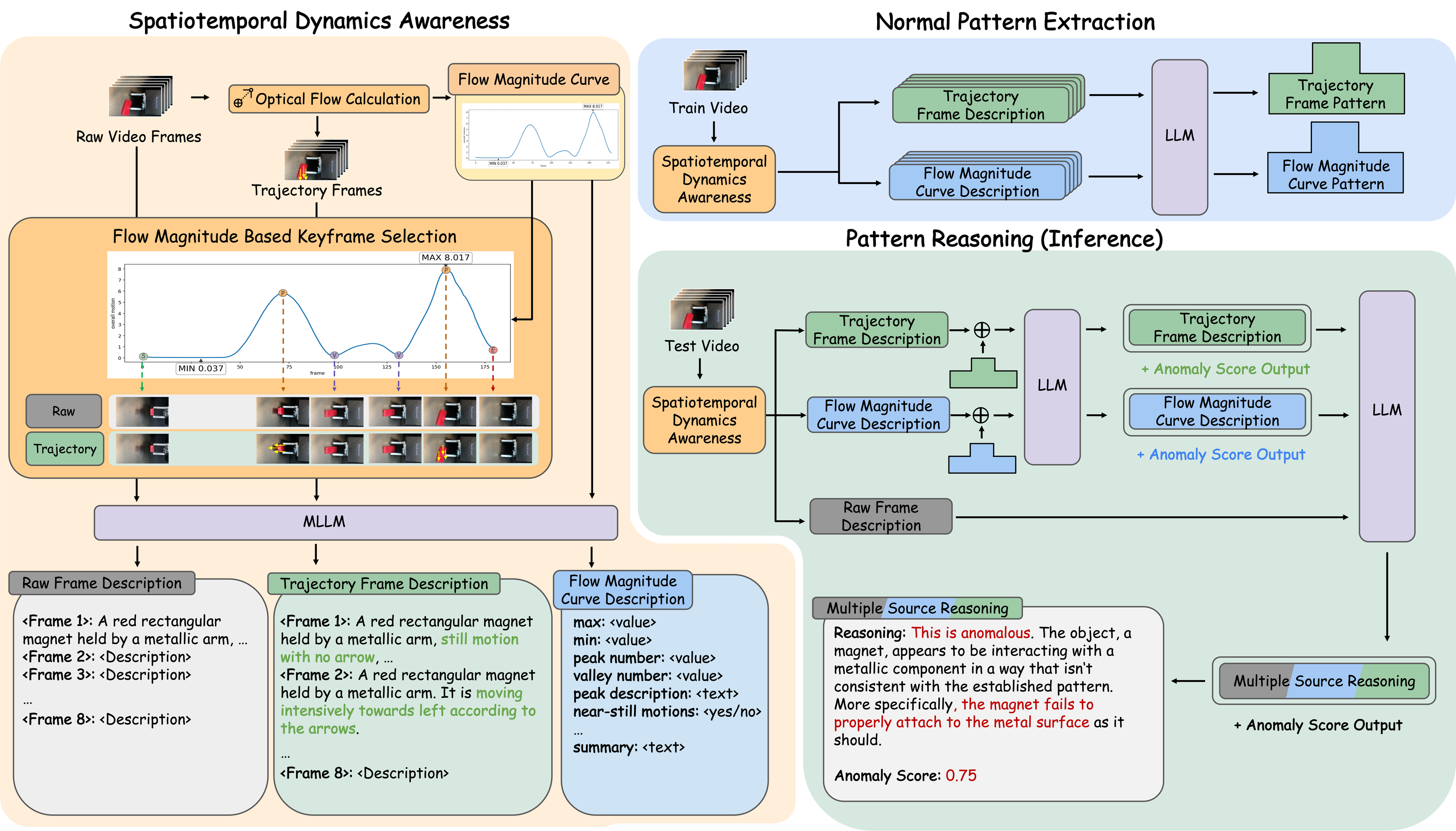}
    \caption{An overview of the proposed MUSTARD framework. It consists of three main stages. (1) Spatiotemporal Dynamics Awareness: Leverages optical flow to select keyframes and derive three complementary descriptive modalities, including raw frames for visual context, trajectory frames for visualizing local motion, and a flow magnitude curve for representing global temporal dynamics. (2) Normal Pattern Extraction: Extracts multi-faceted normal patterns from a small set of reference videos. (3) Pattern Reasoning (Inference): Compares the test video’s description with the extracted normal patterns to detect deviations, producing a final anomaly judgment and explanation.}
    \label{fig:placeholder}
    \vspace{-0.4cm}
\end{figure*}

\section{Related Work}
\label{sec:related_work}

\noindent\textbf{Industrial Anomaly Detection.} 
IAD focuses on products, equipment, and manufacturing processes under controlled conditions. Most existing IAD studies operate at the image level, aiming to detect texture or structural defects in static settings. Typical approaches include autoencoder-based reconstruction (e.g., DRAEM~\cite{zavrtanik2021draem}, UniAD~\cite{you2022unified}, PrincipaLS~\cite{lo2022adversarially}), CLIP-based vision-language alignment~\cite{radford2021learning} (e.g., WinCLIP~\cite{jeong2023winclip}, AdaCLIP~\cite{cao2024adaclip}, AnomalyCLIP~\cite{zhou2024anomalyclip}), and recent MLLM-based frameworks (e.g., AnomalyGPT~\cite{gu2024anomalygpt}, Anomaly-OV~\cite{xu2025towards}, Anomaly-LR~\cite{yeh2026industrial}). However, dynamics-based IAD remains largely unexplored. Specifically, existing approaches rarely incorporate motion continuity, causal reasoning, or temporal cues, which are crucial for detecting dynamic anomalies such as slippage, detachment, or collision. \rev{Recently, Phys-AD~\cite{li2025towards} and IPAD~\cite{ipad} have been introduced as benchmarks for dynamics-based industrial anomalies. Concurrent to our work, O-VAD~\cite{yuan2026vad} uses object-centric state tracking and reasoning, but it is less effective on subtle motion cues and fast state transitions.}

\noindent\textbf{Video Anomaly Detection.} 
Video Anomaly Detection (VAD) targets human-centric surveillance scenes. Existing approaches can be grouped into three main categories: autoencoder-based, CLIP-based, and MLLM-based frameworks. Autoencoder-based methods (e.g., MemAE~\cite{gong2019memorizing}, MNAD~\cite{park2020learning}) learn normal appearance patterns from surveillance videos and detect deviations through reconstruction or prediction errors. CLIP-based methods (e.g., VadCLIP~\cite{vadclip}) leverage vision–language alignment to score anomalous events. Recent MLLM-based frameworks (e.g., LAVAD~\cite{zanella2024harnessing}, AnomalyRuler~\cite{yang2024follow}, HAWK~\cite{tang2024hawk}, Holmes-VAU~\cite{zhang2025holmes}) integrate visual features with LLMs to detect anomalies. These VAD studies primarily focus on human-centric surveillance videos that are largely static, where most anomalous events can be identified from static appearance cues. In contrast, dynamics-based IAD videos are fast-paced and object-centric, requiring finer spatiotemporal granularity. Hence, VAD approaches are not directly applicable.

\noindent\textbf{Multimodal Large Language Models.} 
\rev{MLLMs have recently demonstrated strong general reasoning capabilities across vision--language tasks, with representative model families including LLaVA~\cite{liu2023visual}, Qwen-VL~\cite{bai2025qwen2,bai2025qwen3vl}, InternVL~\cite{wang2025internvl35}, and Seed-VL~\cite{seed2025seed1_5vl}.} However, for video tasks, most models still overlook fine-grained temporal cues, relying on uniformly sampled frames or static summaries (e.g., VideoLLaMA~\cite{zhang-etal-2023-video}, Video-LLaVA~\cite{lin-etal-2024-video}). This results in appearance-centric reasoning that neglects subtle temporal dynamics and causal patterns crucial for dynamics-based IAD. Recent attempts, such as StimuVAR~\cite{guo2025stimuvar} and AKS~\cite{tang2025adaptive}, introduce keyframe selection and token sampling strategies to enhance temporal awareness in MLLMs for video understanding. However, these approaches are designed for generic video tasks and offer limited effectiveness in short, motion-sensitive anomaly scenarios, where fine-grained spatiotemporal cues are essential yet insufficiently explored.

\begin{figure*}
    \centering
    \includegraphics[width=0.8\linewidth]{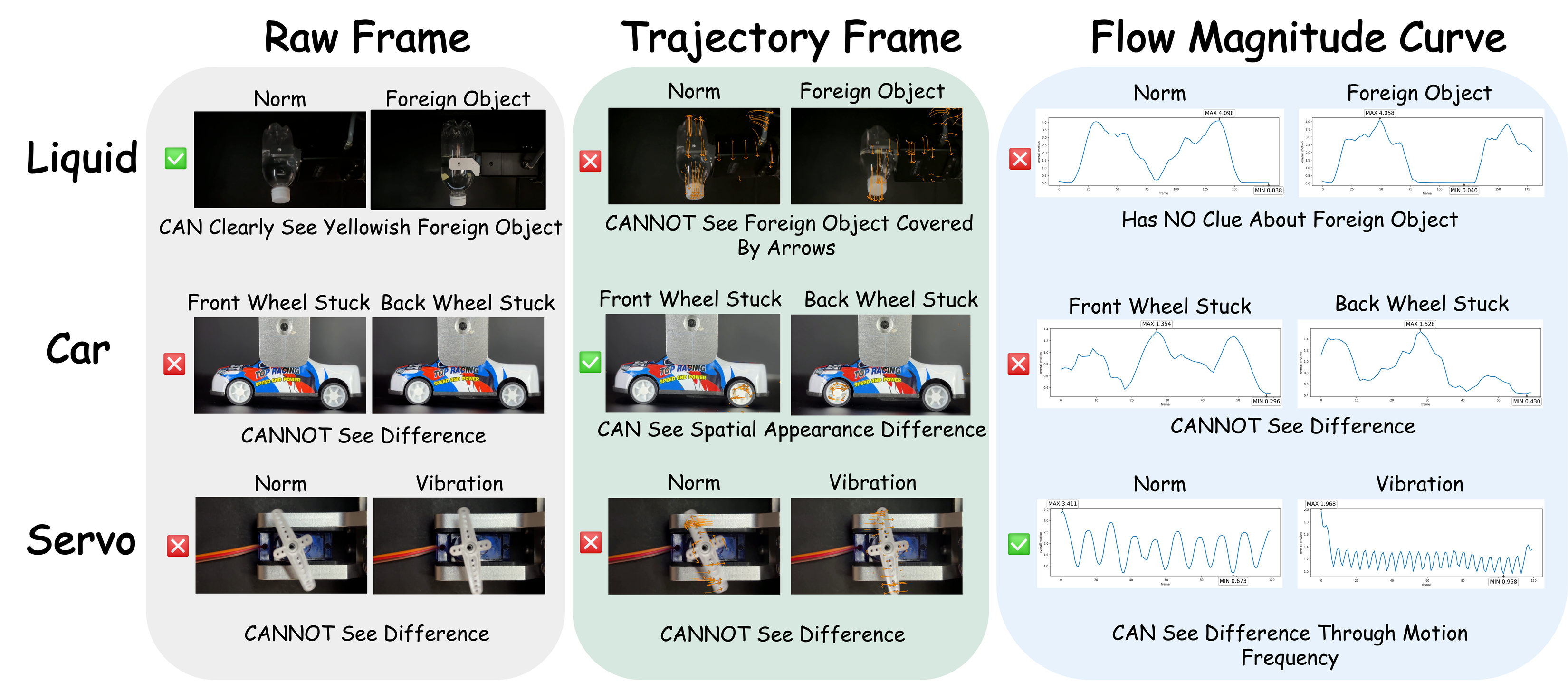}
    \caption{Illustration of the three derived modalities: raw frames, trajectory frames, and the flow magnitude curve. Different sources excel in different categories of test cases. Integrating them enables the model to leverage complementary strengths across modalities.
    }
    \label{fig:source-strength}
    \vspace{-0.4cm}
\end{figure*}

\section{Method}
\label{sec:method}

\subsection{Problem Definition}
We address the dynamics-based IAD problem under the one-class setting. The model is provided with a small number of normal industrial product inspection videos containing dynamic processes. From these normal-only videos, the objective is to learn a comprehensive representation of spatiotemporal normal behavior.

When presented with a previously unseen test video, the model must determine whether it deviates from the learned normal patterns. The output is twofold: (1) a numerical anomaly score and (2) a coherent, human-interpretable natural language explanation that justifies the score. This formulation strictly adheres to the industrial scenario, where anomalous samples are typically unavailable, and both a quantitative score and an interpretable rationale are essential for practical deployment.

\subsection{Framework Overview}
To address the challenge of detecting dynamics-based anomalies, which are often overlooked by existing MLLMs, we propose MUSTARD: Multimodal Spatiotemporal Anomaly Reasoning and Detection, a novel training-free framework that equips models with explicit spatiotemporal awareness. Our core insight is that by translating implicit temporal motion information in videos into explicit textual representations, we enable MLLMs to reason about complex physical dynamics.

As illustrated in Fig.~\ref{fig:placeholder}, the proposed MUSTARD framework consists of three main stages. (1) \textit{Spatiotemporal Dynamics Awareness}. This stage processes a raw input video to extract rich, motion-aware information. It leverages optical flow to select keyframes and derive three complementary descriptive modalities: raw frames for visual context, trajectory frames for visualizing local motion, and a flow magnitude curve for representing global temporal dynamics. (2) \textit{Normal Pattern Extraction}. This stage operates in a few-normal-shot paradigm, extracting multi-faceted normal patterns from a small set of reference videos. (3) \textit{Pattern Reasoning}. During inference, the model compares the test video’s description with the extracted normal patterns to detect deviations, fusing evidence from all three modalities to produce a final anomaly judgment and explanation.

\subsection{Spatiotemporal Dynamics Awareness}
\label{sec:data_preparation}

This stage aims to fully exploit spatiotemporal information in a video to capture fine-grained dynamics-based anomalies. We first approach this from the perspective of global temporal dynamics. Uniform frame sampling is a standard practice in deep learning models, including MLLMs~\cite{zhang-etal-2023-video, lin-etal-2024-video}, for video processing, as it significantly reduces computational cost. However, such uniform sampling overlooks key timesteps that are crucial for detecting abnormal dynamics. To address this limitation, we propose to adopt keyframe selection~\cite{guo2025stimuvar, tang2025adaptive} to capture critical timesteps while maintaining the same number of frames. These critical timesteps are intrinsically well-suited to modeling via optical flow estimation. \rev{Specifically, we employ the Dual TV-L1 optical flow algorithm~\cite{zach2007duality}
to derive pixel-level dense motion between each pair of adjacent frames in the input video.} By aggregating the mean flow magnitudes for each frame, we construct a flow magnitude curve that depicts the intensity of optical flow over time (see Fig.~\ref{fig:placeholder}). Acting as a temporal fingerprint of the video’s overall global dynamics, this curve enables the selection of keyframes corresponding to dynamically significant moments: peaks (maximum motion), valleys (minimum motion), and shoulders (transitional phases). This ensures that the fixed frame sampling budget is allocated to the frames most informative for understanding temporal evolution.


Next, based on the derived optical flow, we construct three complementary modalities for the selected keyframes to further enrich the spatiotemporal context (see Fig.~\ref{fig:source-strength}). (1) \textit{Raw frames}: The original keyframes capture the key ``what'' and ``where'' of the scene. (2) \textit{Trajectory frames}: We capture local motion by overlaying optical-flow directional arrows onto the frames. The arrow size represents motion magnitude, and trailing lines render the motion path history. 
\rev{We initialize uniformly distributed probe points on the first frame and propagate each point through consecutive frames by sampling the local TV-L1 optical-flow vector at its current position, forming continuous motion trajectories.}
To ensure object-centric annotations, we apply the Farthest Point Sampling~\cite{eldar1997farthest} strategy to select a stable and informative subset of points for arrow rendering.(3) \textit{Flow magnitude curve}: The same curve used for keyframe selection, capturing the global temporal dynamics of the entire video. Then, we employ an MLLM to generate text descriptions for each modality, encoding their spatiotemporal information in natural language. The raw frame description focuses on static visual details, such as the main objects, their parts, and their surroundings; the trajectory frame description captures local motion; and the flow magnitude curve description focuses on global temporal dynamics, acting as a plot reader that provides a structured 13-field textual summary (e.g., maximum magnitude, number of peaks, magnitude trends). \rev{The complete modality-description prompts, including the 13-field
flow-magnitude summary format, is provided in Appendix~A7.} Each source exhibits distinct strengths, and we integrate them to complement each other.

\subsection{Normal Pattern Extraction}
\label{sec:reasoning}

The proposed MUSTARD framework adheres to the strict one-class setting via a few-normal-shot paradigm, learning \textit{normal patterns} from a small set of reference normal videos. In our problem, inspection videos may exhibit varying conditions, such as differences in color and operating angle, even for the same product. We therefore aim to extract normal patterns that summarize the reference videos, capturing such variations while aligning descriptions across different videos to form a more robust reference representation.

Specifically, we take a small set of normal training videos as references and apply the Spatiotemporal Dynamics Awareness process to obtain trajectory frame descriptions and flow magnitude curve descriptions for each video. We then adopt a position-wise consolidation strategy to extract robust normal patterns across videos. \rev{First, for each ordered keyframe position, we aggregate descriptions across reference videos and query an LLM to produce a shared pattern. Keyframes at the same position are expected to capture a similar state across videos.} Next, we concatenate these timestep-wise patterns in temporal order and query the LLM to summarize them into global normal patterns, preserving the overall temporal structure of the process. The resulting trajectory frame patterns and flow magnitude curve patterns serve as robust references for detecting dynamics-based anomalies during inference (see Fig.~\ref{fig:placeholder}). \rev{The prompts for normal pattern extraction are provided in Appendix~A7.}

\subsection{Pattern Reasoning}
At inference time, we apply the same Spatiotemporal Dynamics Awareness process to obtain descriptions of the three modalities for each test video. We then perform pattern reasoning via the LLM, where the trajectory and flow magnitude descriptions are compared against their corresponding normal patterns derived from the Normal Pattern Extraction stage. This comparison produces modality-specific anomaly scores along with structured natural language justifications grounded in temporal dynamics.

Crucially, the LLM further performs cross-modal evidence fusion by jointly reasoning over these modality-specific outputs together with the raw frame description (see Fig.~\ref{fig:placeholder}). Through this unified reasoning process, it weighs complementary signals from raw visual context, local motion trajectories, and global temporal dynamics to generate a final anomaly score and coherent explanation. This enables principled dynamics-based anomaly detection with interpretable decision-making. \rev{The prompts for frame descriptions, pattern reasoning, anomaly-score generation, and
cross-modal fusion are provided in Appendix~A7.}

\begin{table*}[t!]
\renewcommand{\arraystretch}{0.75}
\caption{Video-level AUROC (\%) across 22 Phys-AD categories. Scores are scaled by 100, with category-wise best results in \textbf{bold}. Average denotes the unweighted mean across categories.}
\vspace{-0.2cm}
\centering
\label{tab:main_table}
\resizebox{\textwidth}{!}{
\begin{tabular}{l cccc cccc ccc c}
\toprule
\multirow{2}{*}{\textbf{Category}}
& \multicolumn{4}{c}{\textbf{Unsupervised / Weakly-supervised}}
& \multicolumn{4}{c}{\textbf{General-purpose MLLMs}}
& \multicolumn{3}{c}{\textbf{Specialized VAD}}
& \textbf{Ours} \\
\cmidrule(lr){2-5}
\cmidrule(lr){6-9}
\cmidrule(lr){10-12}
\cmidrule(l){13-13}

& MNAD.p & MNAD.r & VADClip & S3R
& V-Chat & V-Llama & V-Llava & Qwen2.5
& LAVAD & ARuler & \rev{O-VAD}
& \textbf{MUSTARD} \\
\midrule

Car
& 49.2 & \textbf{94.4} & 58.1 & 60.6
& 50.0 & 67.8 & 52.2 & 50.0
& 55.7 & 40.7 & \rev{57.5}
& 56.5 \\

Fan
& \textbf{81.0} & 54.2 & 62.4 & 64.0
& 54.9 & 59.2 & 61.1 & 50.0
& 51.0 & 49.4 & \rev{66.9}
& 47.4 \\

Rolling Bearing
& 35.2 & 80.0 & 58.9 & 60.1
& 30.0 & 50.0 & 50.0 & 50.0
& 53.2 & 40.0 & \rev{65.0}
& \textbf{94.2} \\

Spherical Bearing
& \textbf{96.2} & 81.3 & 50.0 & 68.2
& 45.0 & 55.0 & 50.0 & 50.0
& 43.5 & 62.0 & \rev{64.1}
& 76.4 \\

Servo
& \textbf{97.5} & 87.8 & 51.8 & 59.2
& 50.6 & 68.3 & 46.4 & 50.0
& 50.2 & 49.3 & \rev{49.5}
& 69.9 \\

Clip
& 63.0 & 33.3 & 41.2 & 56.1
& 66.9 & 55.6 & 45.8 & 50.0
& 51.6 & 43.8 & \rev{42.3}
& \textbf{86.6} \\

U Disk
& 60.9 & \textbf{94.0} & 53.0 & 54.9
& 56.5 & 57.5 & 50.0 & 50.0
& 51.3 & 46.0 & \rev{60.6}
& 75.3 \\

Hinge
& 70.5 & \textbf{89.5} & 73.7 & 56.1
& 50.0 & 69.2 & 50.0 & 50.0
& 56.4 & 49.9 & \rev{52.6}
& 70.0 \\

Sticky Roller
& 45.1 & \textbf{93.6} & 54.2 & 83.5
& 45.0 & 54.4 & 46.7 & 50.0
& 26.6 & 54.0 & \rev{81.1}
& 68.7 \\

Caster Wheel
& 27.1 & 50.8 & 58.7 & \textbf{67.6}
& 44.4 & 64.2 & 50.0 & 50.0
& 61.5 & 51.2 & \rev{30.1}
& 62.8 \\

Screw
& 68.0 & 54.7 & 50.0 & 65.7
& 47.2 & 25.6 & 55.0 & 50.0
& 68.8 & 47.2 & \rev{70.4}
& \textbf{74.1} \\

Lock
& 43.0 & 64.1 & 45.2 & 52.3
& 27.9 & 49.4 & 50.0 & 50.0
& 34.1 & 47.8 & \rev{78.0}
& \textbf{83.3} \\

Gear
& 65.2 & 69.4 & 51.9 & 58.0
& 54.4 & 60.3 & 51.7 & 50.0
& 60.3 & 55.1 & \rev{\textbf{87.9}}
& 67.7 \\

Clock
& 39.5 & 58.7 & 50.0 & 54.9
& 50.1 & 36.0 & 50.0 & 50.0
& 50.0 & 42.5 & \rev{\textbf{66.9}}
& 55.6 \\

Slide
& \textbf{91.7} & 78.4 & 53.1 & 61.1
& 56.2 & 56.7 & 17.9 & 50.0
& 42.5 & 40.6 & \rev{41.5}
& 70.7 \\

Zipper
& \textbf{82.9} & 42.1 & 50.4 & 63.6
& 54.7 & 48.9 & 50.0 & 49.2
& 53.5 & 46.8 & \rev{38.8}
& 62.7 \\

Button
& \textbf{66.0} & 56.8 & 62.7 & 51.5
& 51.5 & 51.7 & 36.0 & 50.0
& 43.9 & 42.6 & \rev{39.3}
& 62.1 \\

Liquid
& 67.1 & \textbf{83.1} & 54.2 & 59.5
& 41.0 & 27.8 & 21.7 & 53.3
& 50.4 & 77.2 & \rev{77.6}
& 80.9 \\

Rubber Band
& 36.6 & 30.7 & 48.2 & 62.3
& 51.7 & 45.0 & 45.0 & 50.0
& 51.1 & 44.9 & \rev{72.1}
& \textbf{75.3} \\

Ball
& \textbf{72.8} & 68.7 & 50.0 & 66.7
& 56.2 & 63.6 & 53.3 & 50.0
& 60.3 & 47.3 & \rev{57.9}
& 59.1 \\

Magnet
& \textbf{69.1} & 43.8 & 50.0 & 54.8
& 68.3 & 30.0 & 40.0 & 50.0
& 50.2 & 64.6 & \rev{57.6}
& 50.9 \\

Toothpaste
& 52.0 & 63.1 & 50.0 & 68.6
& 37.6 & 55.0 & 46.7 & 50.0
& 56.2 & 53.2 & \rev{\textbf{68.9}}
& 48.7 \\

\midrule
\textbf{Average}
& 62.7 & 66.9 & 53.5 & 61.3
& 49.6 & 52.3 & 46.3 & 50.1
& 51.0 & 49.8 & \rev{60.3}
& \textbf{68.1} \\

\bottomrule
\end{tabular}
}
\vspace{-0.3cm}
\end{table*}
\section{Experiments}
\label{sec:experiments}
\subsection{Experimental Setups}


\noindent\textbf{Datasets.}
Phys-AD~\cite{li2025towards} is the first large-scale, physics-grounded video dataset for industrial anomaly detection, focusing on anomalies revealed through physical interaction and dynamics rather than static defects. The dataset contains over 6,400 videos across 22 object categories (e.g., gears, rubber bands) and 47 anomaly types, requiring models to reason about physical properties like elasticity and friction. To demonstrate the generalization of our method, we also evaluate on the IPAD~\cite{ipad} dataset. It is a specialized industrial benchmark that emphasizes the periodicity of equipment operations. It comprises 597,979 frames across 16 industrial devices, and 39 anomaly classes covering appearance, position, and motion-based failures.

\noindent\textbf{Evaluation Metrics.}
To provide a holistic evaluation of both detection and reasoning, we utilize three metrics:

\begin{itemize}
    \item \textit{Area Under the Receiver Operating Characteristic Curve (AUC)}: The standard, video-level metric for anomaly detection performance.
    
    \item \textit{PAEval}: The Phys-AD dataset introduces an evaluation metric that quantifies the semantic similarity between model anomaly explanation and ground-truth anomaly explanation, serving as a measure of anomaly reasoning quality. PAEval is given ranging from 0 to 1, higher score means better quality of anomaly explanation.
    
    \item \textit{LLM-as-a-Judge}: We evaluated our best method with other methods using LLM-as-a-Judge, where a powerful LLM (GPT-4o~\cite{hurst2024gpt}) is used to pairwise compare two anomaly explanation, based on the ground-truth video descriptions in the Phys-AD dataset.
\end{itemize}

\noindent\textbf{Baseline approaches.}
We compare our method against three categories of baseline approaches:
\begin{itemize}
    \item \textit{Unsupervised and weakly-supervised methods}: MNAD~\cite{park2020learning}, VADClip~\cite{vadclip}, S3R~\cite{S3R} representing traditional VAD methods that only provide anomaly scores without reasoning.
    \item \textit{General-purpose MLLMs}: Video-ChatGPT~\cite{maaz-etal-2024-video}, Video-LLaMA~\cite{zhang-etal-2023-video}, Video-LLaVA~\cite{lin-etal-2024-video}, Qwen2.5-VL~\cite{bai2025qwen2}, as non-specialized MLLMs. Specialized prompts are fed into this model to generate anomaly scores and explanations.
    \item \textit{Specialized VAD MLLMs}: LAVAD~\cite{zanella2024harnessing}, AnomalyRuler~\cite{yang2024follow}, and O-VAD~\cite{yuan2026vad}, as SOTA MLLM-based VAD frameworks.
\end{itemize}

\noindent\textbf{Implementation Details.}
We use Qwen2.5-VL-7B for vision-to-text and Gemma3-12B~\cite{gemmateam2025gemma3technicalreport} for pattern operations. All models are served locally through Ollama. Normal patterns are constructed from $N=30$ videos. For data preparation, we compute dense optical flow using the Dual TV-L1 algorithm. Specific hyperparameters include: we select a fixed budget of 8 keyframes per video based on the flow magnitude curve. For object-centric trajectory annotations, we sample an $80\times50$ grid and rank 600 active points. We select the 10 most active and 40 via Farthest Point Sampling for a total of 50 tracked points. Trajectory trails are rendered over a 12-frame history. 

Performance metrics for unsupervised and weakly-supervised methods, LAVAD and general-purpose MLLMs, with the exception of Qwen2.5-VL-7B which we evaluate directly. For fair comparison with AnomalyRuler, we remove its perception-smoothing module to enable video-level anomaly detection, and employ identical MLLM and LLM models (Qwen2.5-VL-7B and Gemma3-12B), ensuring consistency in backbone models across all experiments.

\subsection{Main Results}

\noindent\textbf{Detection Performance on Phys-AD.}
Table~\ref{tab:main_table} presents the AUC performance of primary anomaly detection frameworks on the Phys-AD dataset. Our method achieves an AUC of 0.681, outperforming all MLLM-based baselines by 0.078 and exceeding the state-of-the-art method by 0.012. Notably, the SOTA approach requires full-shot training and provides no anomaly explanations, whereas our method operates in a few-normal-shot manner without full training while delivering superior anomaly detection performance.
In our architecture, Qwen2.5-VL-7B serves as the primary MLLM backbone for essential video-to-text tasks. However, in Table~\ref{tab:main_table} we can see that directly obtaining video descriptions and querying Qwen2.5-VL-7B to determine if the description is "anomalous or not", is insufficient for accurate detection. This limitation arises because the model’s inherent, general-purpose knowledge of anomalies often fails to align with the highly specific physical and operational constraints of industrial scenarios, highlighting the fundamental difficulty of zero-shot reasoning in this domain. In contrast, our few-normal-shot paradigm leverages normal exemplars to establish domain-specific anomaly criteria, enabling effective detection without extensive training.
Notably, current video anomaly detection (VAD) methods are inferior to handle fine-grained spatiotemporal anomalies in industrial settings. Previous approaches predominantly focus on anomalies detectable within single frames such as surveillance anomaly detection, which fail entirely in physical industrial scenes that require temporal information. Our framework successfully addresses this gap by incorporating flow-annotated frames and flow magnitude curves to capture essential temporal dynamics, enabling accurate detection of motion-dependent anomalies that manifest across multiple frames.

\renewcommand{\arraystretch}{0.75} 
\begin{table}[!t]
\centering
\caption{Generalization Results on IPAD Dataset (AUROC \%)}
\label{tab:ipad_results}
\vspace{-0.2cm}
\begin{tabular}{l|c}
\toprule
\textbf{Method} & \textbf{AUROC $\uparrow$} \\
\midrule
Video-LLaMA~\cite{zhang-etal-2023-video}   & 45.5 \\
Qwen2.5-VL-7B~\cite{bai2025qwen2} & 47.2 \\
Video-ChatGPT~\cite{maaz-etal-2024-video} & 47.9 \\
Video-LLaVA~\cite{lin-etal-2024-video}   & 50.0 \\
\rev{O-VAD~\cite{yuan2026vad}} & \rev{56.5} \\
\midrule
\textbf{Ours (MUSTARD)} & \textbf{57.2} \\
\bottomrule
\end{tabular}
\vspace{-0.5cm}
\end{table}

\noindent\textbf{Generalization to IPAD.}
To further evaluate the robustness of our framework, we test MUSTARD on the IPAD dataset. As shown in Table~\ref{tab:ipad_results}, IPAD presents an exceptionally challenging environment where general-purpose MLLMs often perform near or below the level of random chance ($0.50$). This is likely due to the complex periodicity and subtle motion-based failures inherent in large-scale industrial equipment. Despite these challenges, MUSTARD achieves an AUROC of \textbf{0.572}, surpass all the baselines. This performance confirms that our spatiotemporal awareness module provides a critical signal for detecting anomalies in long-form industrial processes, even when global motion patterns are highly complex.

\noindent\textbf{Anomaly Explanation Performance.}
Table~\ref{tab:reasoning_eval} compares our approach with state-of-the-art MLLM-based video understanding methods on anomaly explanation quality on the Phys-AD dataset. Our method achieves a PAEval score of $0.456$, substantially outperforming all competing methods. This demonstrates the effectiveness of our spatiotemporal reasoning framework. General-purpose video MLLMs and existing MLLM-based anomaly detection methods, such as Anomaly Ruler, obtain considerably lower scores, suggesting that domain-specific design is crucial for anomaly explanation. The large performance gap validates our multi-source spatiotemporal integration, which combines flow-annotated frames, flow magnitude curves, and static descriptions to produce explanations that better align with ground-truth reasoning.

To further evaluate explanation quality, we adopt LLM-as-a-Judge pairwise comparison as a complementary evaluation protocol. Specifically, we use GPT-4o as an expert judge to compare our explanation against each baseline explanation, given the ground-truth frame description. The judge selects the better answer between the two methods, and we report the win rate of other methods versus ours. As shown in Table~\ref{tab:reasoning_eval}, our method achieves win rates substantially above 50\% against all baselines. This confirms that our explanations are not only quantitatively closer to the ground truth under PAEval, but also qualitatively preferred by a strong language model evaluator.

Together, both metrics show that our method provides stronger reasoning ability for video anomaly explanation, establishing a new state of the art for MLLM-based anomaly detection and interpretation.

\begin{figure}[!htbp]
    \centering
    \includegraphics[width=1\linewidth]{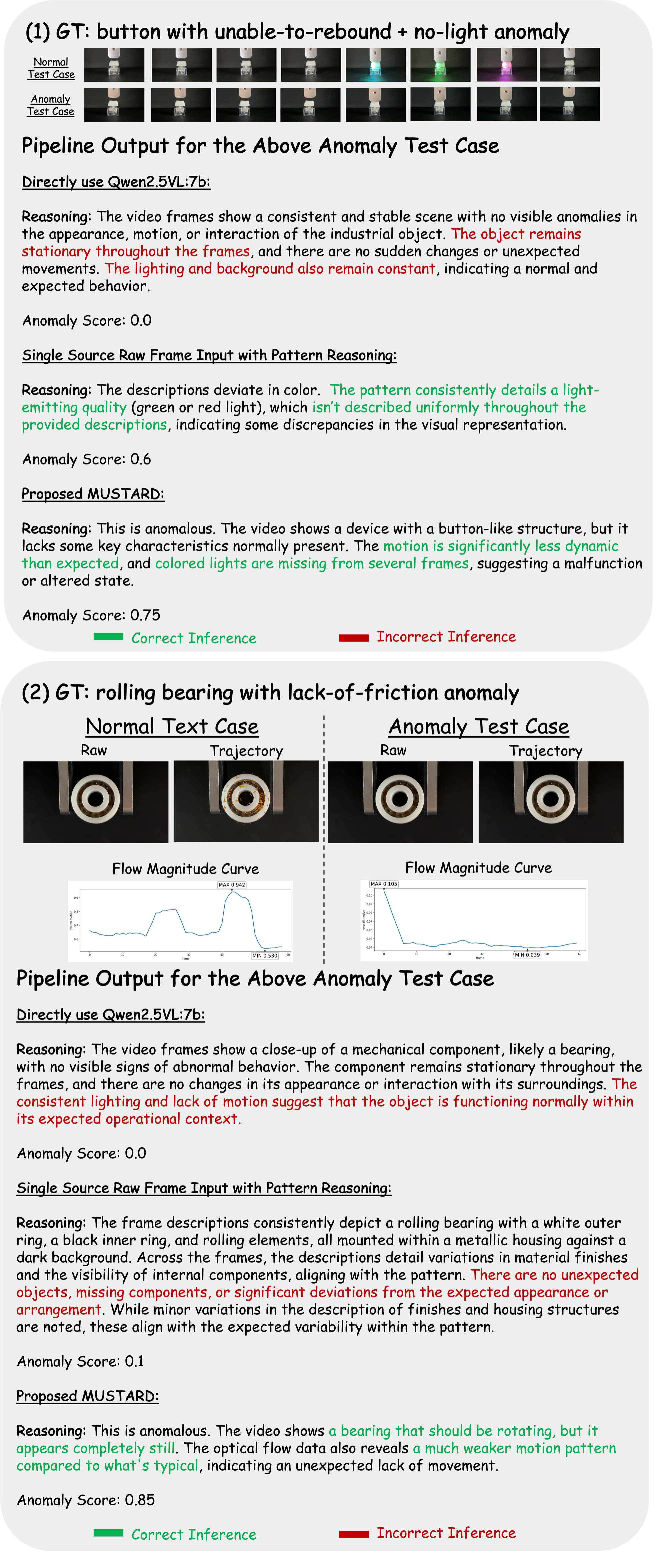}
    \caption{
Example outputs from three methods: direct MLLM usage, a single-source static pattern-reasoning pipeline, and our proposed MUSTARD method.
    }
    \label{fig:example-output}
\end{figure}

\noindent\textbf{Qualitative Results.}
In Figure~\ref{fig:example-output}, we illustrate two representative cases demonstrating how each component of our system contributes to improved anomaly detection. In the first example, direct usage of Qwen2.5-VL correctly recognizes that the button in the scene is not illuminated, but it has no notion of whether this observation is normal or abnormal. Our pattern-reasoning pipeline resolves this ambiguity by comparing the test case's visual description against extracted normal patterns, allowing it to correctly identify the missing illumination as an anomaly. In the second example, both direct Qwen2.5-VL and our static single-source pipeline fail because the anomaly arises from subtle motion behavior rather than appearance. This is precisely where our spatiotemporal dynamics awareness becomes essential. MUSTARD integrates trajectory keyframes and the flow magnitude curve, enabling it to pinpoint that the rolling bearing remains static when it should be rotating. By contrasting these spatiotemporal signals with the normal dynamic patterns, MUSTARD successfully infers that the video is anomalous. These examples jointly demonstrate that our pattern-reasoning pipeline and our motion-aware module address different failure modes, and together enable the full MUSTARD pipeline to robustly detect diverse types of anomalies.



\renewcommand{\arraystretch}{0.75} 
\begin{table}[t!]
\centering
\caption{Anomaly Explanation Quality and Pairwise Win Rates versus Ours (\%). A lower win rate indicates that the baseline is less competitive than our method. All baselines achieve win rates below 50\%, demonstrating the superiority of our method.}
\label{tab:reasoning_eval}
\vspace{-0.2cm}
\begin{tabular}{l|c|c}
\toprule
\textbf{Method} & \textbf{PAEval $\uparrow$} & \textbf{Win Rate vs. Ours} \\
\midrule
Video-LLaMA~\cite{zhang-etal-2023-video}   & 30.3 & 6.81 \\
Video-LLaVA~\cite{lin-etal-2024-video}   & 28.2 & 9.28 \\
Qwen2.5-VL-7B~\cite{bai2025qwen2} & 22.3 & 22.63 \\
AnomalyRuler~\cite{yang2024follow} & 26.3 & 34.65 \\
\midrule
\textbf{Ours (MUSTARD)} & \textbf{45.6} & -- \\
\bottomrule
\end{tabular}
\vspace{-0.5cm}
\end{table}

\subsection{Ablation Study}

We conduct a series of ablation studies to validate our core design choices, isolating the contributions of our flow-guided keyframe selection, spatiotemporal sources.

\noindent\textbf{Flow-guided Keyframe Selection.}
We have done experiment analyzing the impact of our flow-guided keyframe selection strategy against uniform sampling methods. Here, raw video frame pattern reasoning is used to ensure a fair comparison. Standard uniform sampling provides a low baseline AUC of 0.509. Our flow-guided keyframe selection improves this to 0.534 AUC. This demonstrates that flow-guided keyframe selection is capable of choosing the informative frames of objects in the video.

\renewcommand{\arraystretch}{0.75} 
\begin{table}[t]
\centering
\caption{Ablation Study of Spatiotemporal Modalities on Detection and Reasoning. Scores are scaled by 100 for readability.}
\label{tab:ablation_sources}
\vspace{-0.2cm}
\begin{tabular}{ccc|cc}
\toprule
\makecell{\textbf{Raw}\\\textbf{Frames}} &
\makecell{\textbf{Traj.}\\\textbf{Frames}} &
\makecell{\textbf{Flow}\\\textbf{Mag.}} &
\makecell{\textbf{AUROC} $\uparrow$} &
\makecell{\textbf{PAEval} $\uparrow$} \\
\midrule
\checkmark &           &           & 53.4 & 15.79 \\
           & \checkmark &           & 57.0 & 11.20 \\
\checkmark &           & \checkmark & 64.7 & \textbf{45.88} \\
           & \checkmark & \checkmark & 67.1 & 44.63 \\
\midrule
\textbf{\checkmark} & \textbf{\checkmark} & \textbf{\checkmark} & \textbf{68.1} & 45.62 \\
\bottomrule
\end{tabular}
\vspace{-0.5cm}
\end{table}

\noindent\textbf{Spatiotemporal Information Sources.}
Table~\ref{tab:ablation_sources} evaluates the contribution of each spatiotemporal data source in our framework. All variants use our flow-guided keyframe selection strategy for fair comparison.

The raw frame description baseline achieves $0.534$ AUC. Incorporating dynamic information through flow-annotated frames improves performance to $0.570$ AUC, showing that explicit motion visualization enhances the MLLM’s spatiotemporal reasoning. Combining raw frame descriptions with flow magnitude curve pattern reasoning improves AUC to $0.647$, confirming that global motion dynamics provide complementary information for anomaly detection. Our framework, which integrates pattern reasoning from flow-annotated frames and flow magnitude curves while preserving raw frame descriptions for knowledge alignment, achieves the best performance at $0.681$ AUC. This supports our hypothesis that these three sources encode complementary spatiotemporal cues for anomaly identification.

For PAEval, the raw frame description baseline achieves $0.1579$, while flow-annotated frames alone decrease the score to $0.1120$ despite improving AUC. This reveals a trade-off: motion arrows help detection but introduce visual artifacts that may impair explanation quality. Adding flow magnitude curves to raw frames yields a substantial improvement to $0.4588$, nearly tripling the baseline score. Replacing raw frames with flow-annotated frames slightly reduces PAEval to $0.4463$, further indicating the noise introduced by motion annotations. Our full framework combines \rev{raw descriptions} alongside both dynamic sources, recovers PAEval to $0.4562$, effectively matching the raw-frame-plus-curve setting while achieving superior AUC.

This ablation highlights three findings: (1) flow magnitude curves provide the largest boost for both detection and explanation, (2) flow-annotated frames improve detection but must be integrated carefully to avoid degrading explanations, and (3) our multi-source fusion strategy effectively leverages all three complementary modalities by balancing their strengths and mitigating their weaknesses.

\section{Conclusion}
\label{sec:conlcusion}
We introduce MUSTARD, a novel framework that equips MLLMs with optical flow-guided spatiotemporal awareness for IAD. By using optical flow to select keyframes capturing critical motion irregularities and incorporating two flow-augmented modalities, MUSTARD significantly enhances fine-grained temporal anomaly identification. To our knowledge, MUSTARD is one of the first MLLM-based frameworks for dynamics-based IAD. Our experiments demonstrate SOTA performance in spatiotemporal anomaly detection with interpretable explanations. 

\section*{Acknowledgements}
This work was supported by the National Science and Technology Council (NSTC) of Taiwan, under Grant 115-2634-F-002-012, and by the NTU AI Center of Research Excellence (AI-CoRE). Shao-Yuan Lo is honored to be supported by the Yushan Young Fellow Program by the Ministry of Education (MOE) of Taiwan.

{
    \small
    \bibliographystyle{ieeenat_fullname}
    \bibliography{main,supp}

@String(CVPR= {IEEE Conf. Comput. Vis. Pattern Recog.})

@String(AAAI = {AAAI})

@String(CVPR  = {CVPR})

@inproceedings{li2025towards,
  title={Towards visual discrimination and reasoning of real-world physical dynamics: Physics-grounded anomaly detection},
  author={Li, Wenqiao and Gu, Yao and Chen, Xintao and Xu, Xiaohao and Hu, Ming and Huang, Xiaonan and Wu, Yingna},
  booktitle={IEEE/CVF Conference on Computer Vision and Pattern Recognition},
  year={2025}
}

@inproceedings{yang2024follow,
  title={Follow the rules: Reasoning for video anomaly detection with large language models},
  author={Yang, Yuchen and Lee, Kwonjoon and Dariush, Behzad and Cao, Yinzhi and Lo, Shao-Yuan},
  booktitle={European Conference on Computer Vision},
  year={2024}
}

@inproceedings{guo2025stimuvar,
  title={Stimuvar: Spatiotemporal stimuli-aware video affective reasoning with multimodal large language models},
  author={Guo, Yuxiang and Siddiqui, Faizan and Zhao, Yang and Chellappa, Rama and Lo, Shao-Yuan},
  booktitle={International Journal of Computer Vision},
  year={2025}
}

@inproceedings{gong2019memorizing,
  title={Memorizing normality to detect anomaly: Memory-augmented deep autoencoder for unsupervised anomaly detection},
  author={Gong, Dong and Liu, Lingqiao and Le, Vuong and Saha, Budhaditya and Mansour, Moussa Reda and Venkatesh, Svetha and Hengel, Anton van den},
  booktitle={IEEE/CVF International Conference on Computer Vision},
  year={2019}
}

@inproceedings{park2020learning,
  title={Learning memory-guided normality for anomaly detection},
  author={Park, Hyunjong and Noh, Jongyoun and Ham, Bumsub},
  booktitle={IEEE/CVF Conference on Computer Vision and Pattern Recognition},
  year={2020}
}

@inproceedings{vadclip,
  title={Vadclip: Adapting vision-language models for weakly supervised video anomaly detection},
  author={Wu, Peng and Zhou, Xuerong and Pang, Guansong and Zhou, Lingru and Yan, Qingsen and Wang, Peng and Zhang, Yanning},
  booktitle={AAAI Conference on Artificial Intelligence},
  year={2024}
}

@inproceedings{S3R,
  title={Self-supervised sparse representation for video anomaly detection},
  author={Wu, Jhih-Ciang and Hsieh, He-Yen and Chen, Ding-Jie and Fuh, Chiou-Shann and Liu, Tyng-Luh},
  booktitle={European Conference on Computer Vision},
  year={2022}
}

@inproceedings{zanella2024harnessing,
  title={Harnessing Large Language Models for Training-free Video Anomaly Detection},
  author={Zanella, Luca and Menapace, Willi and Mancini, Massimiliano and Wang, Yiming and Ricci, Elisa},
  booktitle={IEEE/CVF Conference on Computer Vision and Pattern Recognition},
  year={2024}
}

@inproceedings{radford2021learning,
  title={Learning transferable visual models from natural language supervision},
  author={Radford, Alec and Kim, Jong Wook and Hallacy, Chris and Ramesh, Aditya and Goh, Gabriel and Agarwal, Sandhini and Sastry, Girish and Askell, Amanda and Mishkin, Pamela and Clark, Jack and others},
  booktitle={International Conference on Machine Learning},
  year={2021}
}

@inproceedings{maaz-etal-2024-video,
    title = "Video-{C}hat{GPT}: Towards Detailed Video Understanding via Large Vision and Language Models",
    author = "Maaz, Muhammad  and
      Rasheed, Hanoona  and
      Khan, Salman  and
      Khan, Fahad",
    booktitle = "ACL",
    year = "2024"
}

@inproceedings{zhang-etal-2023-video,
    title = "Video-{LL}a{MA}: An Instruction-tuned Audio-Visual Language Model for Video Understanding",
    author = "Zhang, Hang  and
      Li, Xin  and
      Bing, Lidong",
    editor = "Feng, Yansong  and
      Lefever, Els",
    booktitle = "Conference on Empirical Methods in Natural Language Processing",
    year = "2023"
}

@inproceedings{lin-etal-2024-video,
    title = "Video-{LL}a{VA}: Learning United Visual Representation by Alignment Before Projection",
    author = "Lin, Bin  and
      Ye, Yang  and
      Zhu, Bin  and
      Cui, Jiaxi  and
      Ning, Munan  and
      Jin, Peng  and
      Yuan, Li",
    editor = "Al-Onaizan, Yaser  and
      Bansal, Mohit  and
      Chen, Yun-Nung",
    booktitle = "Conference on Empirical Methods in Natural Language Processing",
    year = "2024"
}

@inproceedings{liu2023visual,
  title={Visual instruction tuning},
  author={Liu, Haotian and Li, Chunyuan and Wu, Qingyang and Lee, Yong Jae},
  booktitle={Conference on Neural Information Processing Systems},
  year={2023}
}

@inproceedings{tang2025adaptive,
  title={Adaptive keyframe sampling for long video understanding},
  author={Tang, Xi and Qiu, Jihao and Xie, Lingxi and Tian, Yunjie and Jiao, Jianbin and Ye, Qixiang},
  booktitle={IEEE/CVF Conference on Computer Vision and Pattern Recognition},
  year={2025}
}

@inproceedings{tang2024hawk,
  title={Hawk: Learning to understand open-world video anomalies},
  author={Tang, Jiaqi and Lu, Hao and Wu, Ruizheng and Xu, Xiaogang and Ma, Ke and Fang, Cheng and Guo, Bin and Lu, Jiangbo and Chen, Qifeng and Chen, Yingcong},
  booktitle={Conference on Neural Information Processing Systems},
  year={2024}
}

@inproceedings{zhang2025holmes,
  title={Holmes-vau: Towards long-term video anomaly understanding at any granularity},
  author={Zhang, Huaxin and Xu, Xiaohao and Wang, Xiang and Zuo, Jialong and Huang, Xiaonan and Gao, Changxin and Zhang, Shanjun and Yu, Li and Sang, Nong},
  booktitle={IEEE/CVF Conference on Computer Vision and Pattern Recognition},
  year={2025}
}

@inproceedings{jeong2023winclip,
  title={Winclip: Zero-/few-shot anomaly classification and segmentation},
  author={Jeong, Jongheon and Zou, Yang and Kim, Taewan and Zhang, Dongqing and Ravichandran, Avinash and Dabeer, Onkar},
  booktitle={CVPR},
  year={2023}
}

@inproceedings{cao2024adaclip,
  title={Adaclip: Adapting clip with hybrid learnable prompts for zero-shot anomaly detection},
  author={Cao, Yunkang and Zhang, Jiangning and Frittoli, Luca and Cheng, Yuqi and Shen, Weiming and Boracchi, Giacomo},
  booktitle={European Conference on Computer Vision},
  year={2024}
}

@inproceedings{gu2024anomalygpt,
  title={Anomalygpt: Detecting industrial anomalies using large vision-language models},
  author={Gu, Zhaopeng and Zhu, Bingke and Zhu, Guibo and Chen, Yingying and Tang, Ming and Wang, Jinqiao},
  booktitle={AAAI Conference on Artificial Intelligence},
  year={2024}
}

@inproceedings{xu2025towards,
  title={Towards zero-shot anomaly detection and reasoning with multimodal large language models},
  author={Xu, Jiacong and Lo, Shao-Yuan and Safaei, Bardia and Patel, Vishal M and Dwivedi, Isht},
  booktitle={IEEE/CVF Conference on Computer Vision and Pattern Recognition},
  year={2025}
}

@inproceedings{
zhou2024anomalyclip,
    title={Anomaly{CLIP}: Object-agnostic Prompt Learning for Zero-shot Anomaly Detection},
    author={Qihang Zhou and Guansong Pang and Yu Tian and Shibo He and Jiming Chen},
    booktitle={International Conference on Learning Representations},
    year={2024}
}

@inproceedings{zach2007duality,
  author    = {Zach, Christopher and Pock, Thomas and Bischof, Horst},
  title     = {A Duality Based Approach for Realtime TV-L1 Optical Flow},
  booktitle = {Pattern Recognition},
  year      = {2007},
  pages     = {214--223},
  publisher = {Springer Berlin Heidelberg}
}

@inproceedings{bergmann2019mvtec,
  title={MVTec AD--A comprehensive real-world dataset for unsupervised anomaly detection},
  author={Bergmann, Paul and Fauser, Michael and Sattlegger, David and Steger, Carsten},
  booktitle={IEEE/CVF Conference on Computer Vision and Pattern Recognition},
  year={2019}
}

@article{heckler2025mvtec,
  title={The mvtec ad 2 dataset: Advanced scenarios for unsupervised anomaly detection},
  author={Heckler-Kram, Lars and Neudeck, Jan-Hendrik and Scheler, Ulla and K{\"o}nig, Rebecca and Steger, Carsten},
  journal={arXiv preprint arXiv:2503.21622},
  year={2025}
}

@inproceedings{zou2022spot,
author = {Zou, Yang and Jeong, Jongheon and Pemula, Latha and Zhang, Dongqing and Dabeer, Onkar},
title = {SPot-the-Difference Self-supervised Pre-training for Anomaly Detection and Segmentation},
year = {2022},
booktitle = {European Conference on Computer Vision}
}

@article{bergmann2022beyond,
  title={Beyond dents and scratches: Logical constraints in unsupervised anomaly detection and localization},
  author={Bergmann, Paul and Batzner, Kilian and Fauser, Michael and Sattlegger, David and Steger, Carsten},
  journal={International Journal of Computer Vision},
  volume={130},
  number={4},
  pages={947--969},
  year={2022},
  publisher={Springer}
}

@article{bai2025qwen2,
  title={Qwen2. 5-vl technical report},
  author={Bai, Shuai and Chen, Keqin and Liu, Xuejing and Wang, Jialin and Ge, Wenbin and Song, Sibo and Dang, Kai and Wang, Peng and Wang, Shijie and Tang, Jun and others},
  journal={arXiv preprint arXiv:2502.13923},
  year={2025}
}

@ARTICLE{ipad,
  author={Liu, Jinfan and Yan, Yichao and Li, Junjie and Zhao, Weiming and Chu, Pengzhi and Sheng, Xingdong and Liu, Yunhui and Yang, Xiaokang},
  journal={IEEE Transactions on Circuits and Systems for Video Technology}, 
  title={IPAD: Industrial Process Anomaly Detection Dataset}, 
  year={2025}
}

@inproceedings{zavrtanik2021draem,
  title={Draem-a discriminatively trained reconstruction embedding for surface anomaly detection},
  author={Zavrtanik, Vitjan and Kristan, Matej and Sko{\v{c}}aj, Danijel},
  booktitle={IEEE/CVF International Conference on Computer Vision},
  year={2021}
}

@inproceedings{you2022unified,
  title={A unified model for multi-class anomaly detection},
  author={You, Zhiyuan and Cui, Lei and Shen, Yujun and Yang, Kai and Lu, Xin and Zheng, Yu and Le, Xinyi},
  booktitle={Conference on Neural Information Processing Systems},
  year={2022}
}

@article{lo2022adversarially,
  title={Adversarially robust one-class novelty detection},
  author={Lo, Shao-Yuan and Oza, Poojan and Patel, Vishal M},
  journal={IEEE Transactions on Pattern Analysis and Machine Intelligence},
  year={2022}
}

@inproceedings{wang2024real,
  title={Real-iad: A real-world multi-view dataset for benchmarking versatile industrial anomaly detection},
  author={Wang, Chengjie and Zhu, Wenbing and Gao, Bin-Bin and Gan, Zhenye and Zhang, Jiangning and Gu, Zhihao and Qian, Shuguang and Chen, Mingang and Ma, Lizhuang},
  booktitle={IEEE/CVF Conference on Computer Vision and Pattern Recognition},
  year={2024}
}

@inproceedings{eldar1997farthest,
  title={The farthest point strategy for progressive image sampling},
  author={Eldar, Yuval and Lindenbaum, Michael and Porat, Moshe and Zeevi, Yehoshua Y},
  booktitle={IEEE Transactions on Image Processing},
  year={1997}
}

@article{bai2025qwen3vl,
  title   = {Qwen3-VL Technical Report},
  author  = {Bai, Shuai and others},
  journal = {arXiv preprint arXiv:2511.21631},
  year    = {2025}
}

@article{wang2025internvl35,
  title   = {InternVL3.5: Advancing Open-Source Multimodal Models in Versatility, Reasoning, and Efficiency},
  author  = {Wang, Weiyun and others},
  journal = {arXiv preprint arXiv:2508.18265},
  year    = {2025}
}

@article{seed2025seed1_5vl,
  title   = {Seed1.5-VL Technical Report},
  author  = {{ByteDance Seed Team}},
  journal = {arXiv preprint arXiv:2505.07062},
  year    = {2025}
}

@article{gemmateam2025gemma3technicalreport,
      title={Gemma 3 Technical Report}, 
      author={Gemma Team},
      year={2025},
      journal={arXiv preprint arXiv:2503.19786},
}

@article{hurst2024gpt,
      title={Gpt-4o system card}, 
      author={Hurst, Aaron and Lerer, Adam and Goucher, Adam P and Perelman, Adam and Ramesh, Aditya and Clark, Aidan and Ostrow, AJ and Welihinda, Akila and Hayes, Alan and Radford, Alec and others},
      year={2024},
      journal={arXiv preprint arXiv:2410.21276},
}

@inproceedings{yuan2026vad,
  title={O-VAD: Industrial Video Anomaly Detection Through Object-Centric Tracking and Reasoning},
  author={Yuan, Mei and Long, Qi and Wu, Qifeng and Li, Zhenyang and Zhao, Yizhou and Wang, Lei and Liu, Yang and Xu, Min},
  booktitle={European Conference on Computer Vision},
  pages={285--302},
  year={2026},
  organization={Springer}
}

@article{yeh2026industrial,
  title={Industrial Anomaly Detection via Defect-Grounded Reasoning in Visual Latent Space},
  author={Yeh, Jaron and Chang, Yen-Wei and Liu, Jiang and Lo, Shao-Yuan},
  journal={arXiv preprint arXiv:2609.29457},
  year={2026}
}

@article{liu2024aligning,
  title={Aligning with human judgement: The role of pairwise preference in large language model evaluators},
  author={Liu, Yinhong and Zhou, Han and Guo, Zhijiang and Shareghi, Ehsan and Vuli{\'c}, Ivan and Korhonen, Anna and Collier, Nigel},
  journal={arXiv preprint arXiv:2403.16950},
  year={2024}
}

@inproceedings{wang2024large,
  title={Large language models are not fair evaluators},
  author={Wang, Peiyi and Li, Lei and Chen, Liang and Cai, Zefan and Zhu, Dawei and Lin, Binghuai and Cao, Yunbo and Kong, Lingpeng and Liu, Qi and Liu, Tianyu and others},
  booktitle={Proceedings of the 62nd Annual Meeting of the Association for Computational Linguistics (Volume 1: Long Papers)},
  pages={9440--9450},
  year={2024}
}
}

\clearpage
\maketitlesupplementary

\setcounter{section}{0}
\renewcommand{\thesection}{A\arabic{section}}

\renewcommand{\theHsection}{supplementary.\arabic{section}}
\renewcommand{\theHsubsection}{supplementary.\arabic{section}.\arabic{subsection}}

\section{Limitations and Potential Negative Societal Impacts}

\paragraph{Limitations.}
Consistent with prevalent MLLM research, our framework assumes the underlying backbones possess sufficient competence. Using weaker models can inevitably compromise overall effectiveness. For instance, models may exhibit hallucinations or struggle to accurately describe fine-grained object states (see Sec.~\ref{sec:qualitative-discussion}). Further examining these backbone dependencies and their impact offers a compelling avenue for future study.

Textual descriptions provide a compact way to compare
test videos with normal references while avoiding the
higher cost of direct visual-feature matching.
Despite potential description errors, MUSTARD still
outperforms vanilla MLLM baselines. Nevertheless,
its performance remains dependent on the perceptual
capabilities of the underlying MLLM.
Incorporating visual features as complementary evidence
is a promising direction for future work.

\paragraph{Potential Negative Social Impact.}
The proposed method involves detailed motion analysis, which may enable malicious actors to adapt the system for unauthorized surveillance. Additionally, the reliance on VLMs/LLMs contributes to general environmental carbon footprints. To mitigate these risks, security mechanisms and energy-efficient inference techniques could be integrated.

\section{Evaluation metrics details}

\paragraph{AUC}
For non-LLM-based methods, the framework directly outputs an anomaly score. For LLM-based methods, we prompt the LLM to generate both an anomaly score (ranging from 0 to 1, where 0 indicates completely normal and 1 indicates completely anomalous) and the anomaly explanation. We extract the anomaly score for AUC evaluation and the anomaly explanation for PAEval and LLM-as-a-Judge metrics.

\paragraph{PAEval}

PAEval is the metric proposed in the Phys-AD ~\cite{li2025towards}. The original metric contains two sub-metric: video descriptions and anomaly explanations; we concentrate on anomaly explanations because they provide a direct assessment of a model’s ability to capture the reasoning rationale underlying anomaly detection. Each video has 5 ground-truth anomaly explanations. PAEval measures semantic text similarity between test anomaly explanations and ground-truth anomaly explanations, yielding a score between 0 and 1, and higher score reflects higher similarity, and therefore better anomaly explanations. PAEval selects the highest similarity score across all 5 ground-truth anomaly explanations for each test video. We extract the explanation segment from the LLM output and compute similarity using GPT-4o (2024-08-06), consistent with the Phys-AD dataset evaluation protocol.

\paragraph{LLM-as-a-Judge}

We employ LLM-as-a-Judge with pairwise comparison for each anomaly type in all objects, as LLMs demonstrate greater robustness in comparative judgments than absolute scoring ~\cite{liu2024aligning} in PAEval. The judge LLM receives the ground-truth video description from the PAEval metric (though the metric is not used here) and two candidate anomaly explanations, then selects the anomaly explanation that better captures the anomaly. Win rate is calculated as the proportion of instances where one method's explanation is preferred over another.
We use GPT-4o (2024-08-06) as the judge LLM and evaluate 5 representative videos per anomaly type. To mitigate positional bias ~\cite{wang2024large}, we randomly swap the order of the two candidate explanations in each comparison.

\section{Qualitative Discussion}
\label{sec:qualitative-discussion}

\subsection{Data Source}

We provide a qualitative analysis of our MUSTARD method across various anomaly types to demonstrate the effectiveness of our proposed multi-source data representations. Our approach leverages three complementary data sources: raw frame descriptions (direct visual observations), trajectory frame descriptions (motion-enhanced visualizations with overlaid arrows), and flow-mag curves (optical flow magnitude curves capturing global motion patterns). While initial experiments with raw frames alone yielded suboptimal results, we recognize their fundamental importance and strategically integrate all three sources in MUSTARD to enhance reasoning capabilities.

\subsubsection*{Success Cases}

We first examine scenarios where the integration of trajectory and flow-mag data successfully compensates for visual ambiguity or enhances detection confidence.

\begin{figure*}[!p]
    \centering
    \includegraphics[width=0.85\linewidth]{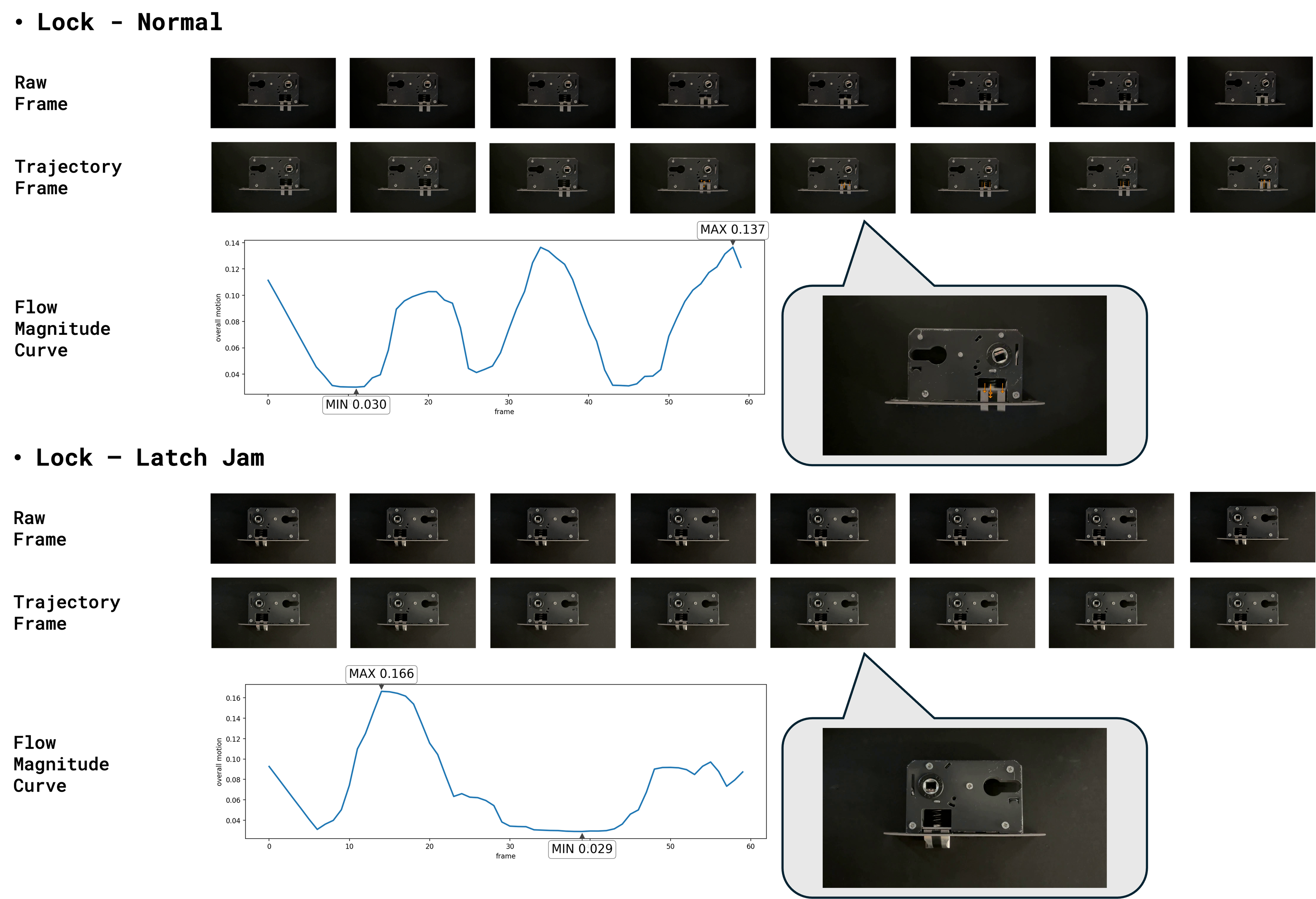}
    \caption{An example of three different data sources for the Lock object.}
    \label{fig:lock}
\end{figure*}

\paragraph{Rolling Bearing and Lock.} These cases (Figure~\ref{fig:rolling_bearing} and Figure~\ref{fig:lock}) exemplify the strength of trajectory frame descriptions for motion-based anomalies. Normal videos display consistent arrow patterns indicating smooth mechanical operations, while abnormal videos show irregular or absent arrows reflecting mechanical failures or jamming. The trajectory descriptions provide clear visual signatures that enable reliable anomaly detection, demonstrating how explicit motion encoding helps the model identify anomalies that are subtle in raw frames alone.

\begin{figure*}[!p]
    \centering
    \includegraphics[width=0.85\linewidth]{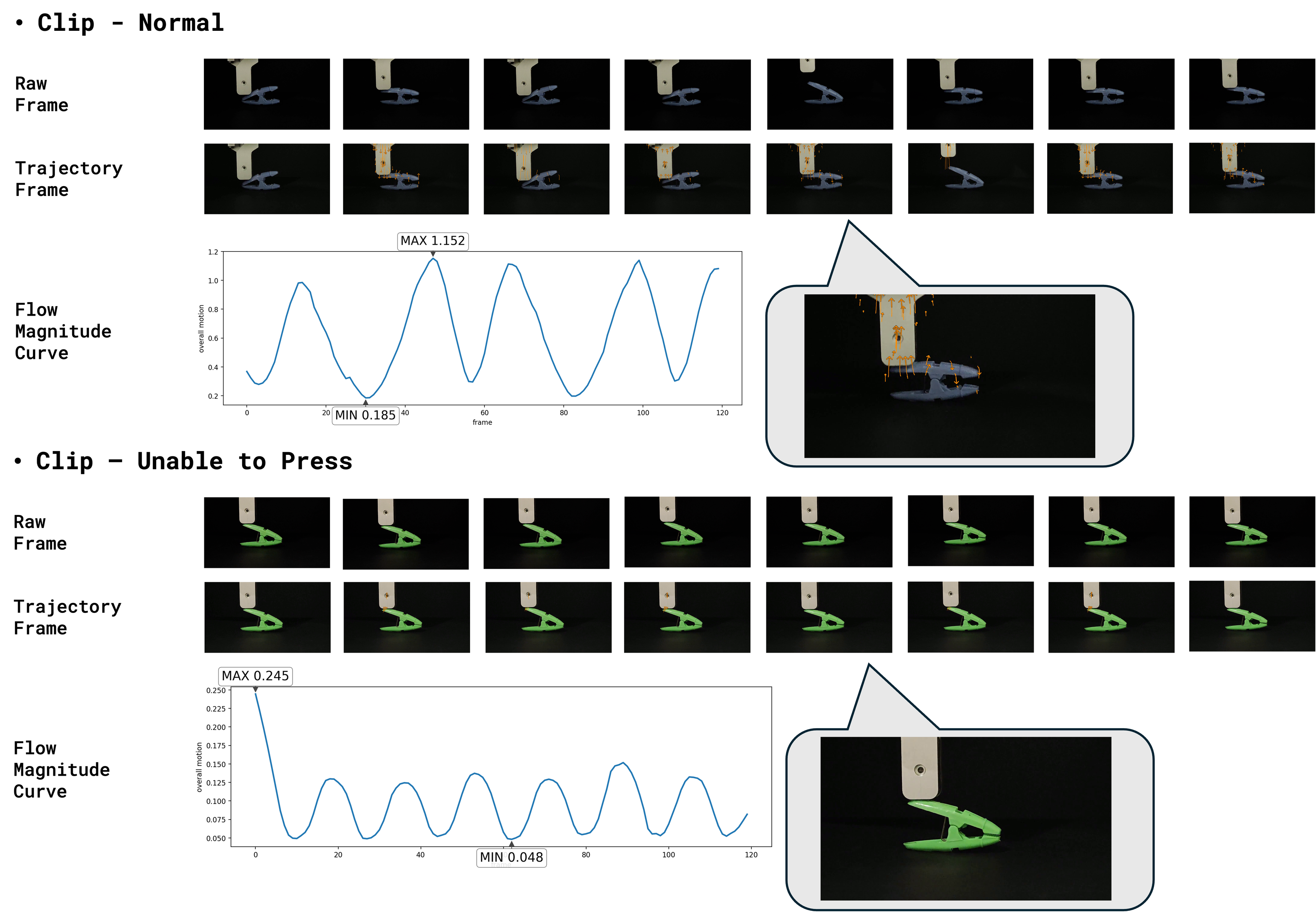}
    \caption{An example of three different data sources for the Clip object.}
    \label{fig:clip}
\end{figure*}

\paragraph{Clip.} This scenario (Figure~\ref{fig:clip}) highlights the superiority of flow-mag curves when frame-based analysis proves insufficient. Although visual appearance and local motion patterns are similar between normal and abnormal clips, the optical flow magnitude curves reveal distinctive global motion signatures. Normal clips exhibit characteristic flow-mag patterns corresponding to proper mechanical action, while abnormal clips display measurably different magnitude profiles. This explicitly validates our motivation for introducing flow-mag curves as a complementary data source that captures motion information at a holistic level.


\subsubsection*{Challenging Cases}

In contrast, the following cases highlight scenarios where motion cues are minimal, forcing reliance on the specific backbone's visual encoding capabilities.

\paragraph{Toothpaste.} The anomaly manifests as a small toothpaste leak, which is a subtle visual change requiring fine-grained observation. Motion patterns remain nearly identical between normal and abnormal videos, rendering trajectory descriptions and flow-mag curves ineffective. Consequently, detection relies entirely on the backbone’s raw visual perception. In this scenario, our selected backbone, Qwen2.5-VL-7B, struggles to notice minor discrepancies, exposing the intrinsic limitations regarding micro-level detail extraction in smaller-scale models compared to larger SOTA LLMs.

\paragraph{Ball.} The anomaly involves ball deformation or deflation after compression by robotic arms. While more substantial than the toothpaste leak, this visual change remains challenging for the selected 7B-parameter backbone to perceive accurately. Since trajectory frame description pattern and flow-mag curves pattern are similar across normal and anomaly videos, both method can offer limited assistance. Detection again depends heavily on the naked visual perception capabilities of the model. The failure here indicates that the perceptual bottleneck of the specific Qwen2.5-VL-7B backbone extends beyond micro-level details to medium-scale visual changes when relying solely on standard visual encoders.

\subsubsection*{Summary}

Our success cases of rolling bearing, lock, and clip, demonstrate the considerable strength of our carefully crafted multi-source data representations. By designing trajectory frame descriptions and flow-mag curves specifically to address motion-based anomalies, we have created a rich, complementary information set that significantly enhances anomaly detection. These data sources effectively encode motion at different scales and perspectives, enabling sophisticated reasoning about physical dynamics.

The challenging cases reveal a specific constraint related to the implementation choice rather than the methodology itself. We utilized a "naked" VLM backbone (Qwen2.5-VL-7B) without applying specialized methods or visual adapters often proposed to enhance visual perception. Consequently, the performance bottleneck reflects the capacity of this specific parameter-efficient backbone to perceive fine-grained details and medium-scale changes when motion cues are insufficient. This suggests that the limitations observed are not inherent to the MUSTARD approach, but rather a consequence of the raw perception capabilities of the selected 7B model. As we adopted a general-purpose, smaller-scale model for this study, future improvements could involve integrating stronger VLM backbones or specialized visual perception modules. As the underlying VLM capabilities scale, we anticipate that the rich, multi-source representations we have created will become even more effective, fully realizing their potential for comprehensive industrial video anomaly detection.



\subsection{Normal Pattern}

\begin{figure*}[!p]
    \centering
    \includegraphics[width=0.85\linewidth]{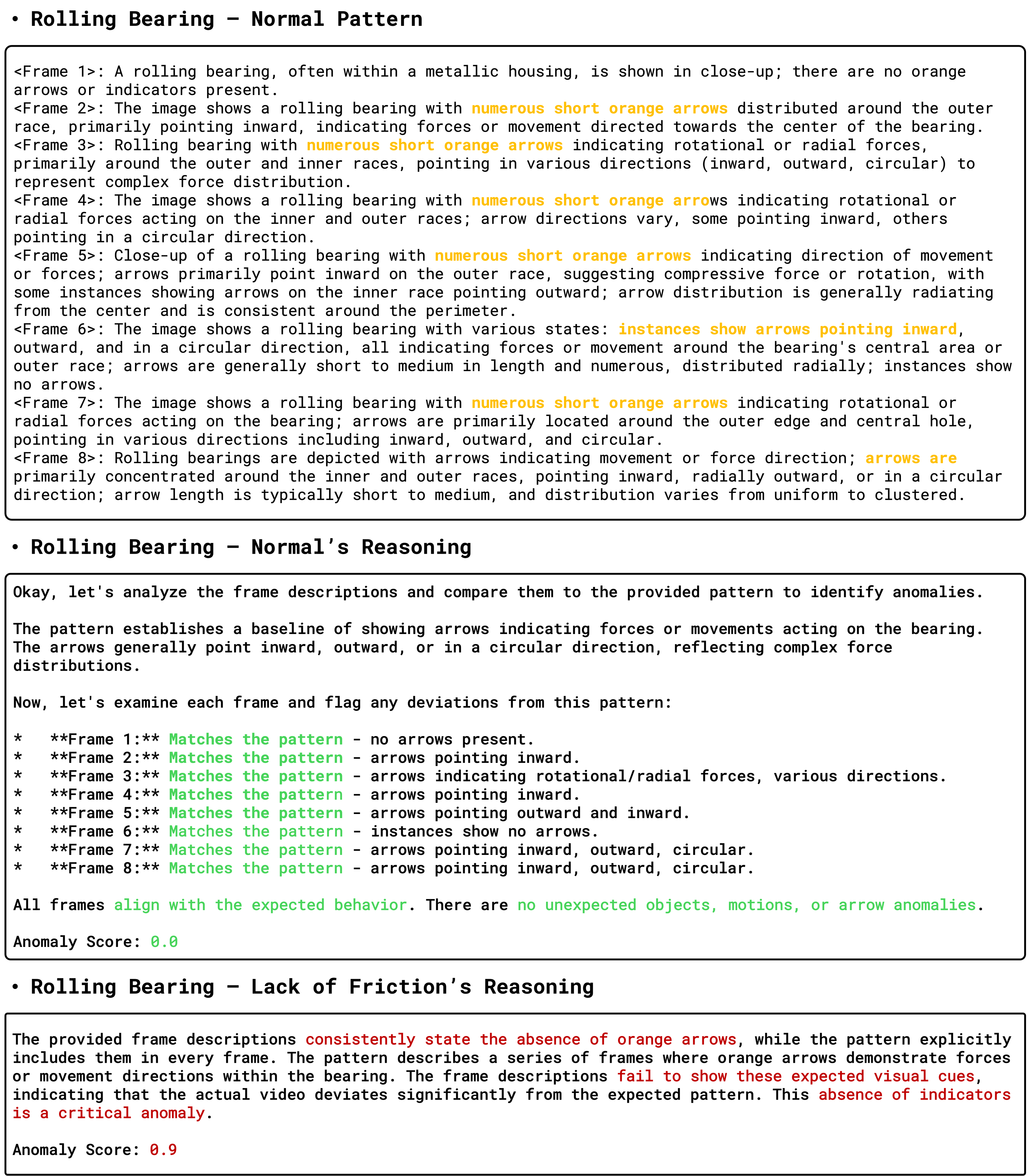}
    \caption{An example of the trajectory-frame normal pattern and corresponding reasoning output.}
    \label{fig:dynamic_pattern}
\end{figure*}

\begin{figure*}[!p]
    \centering
    \includegraphics[width=0.65\linewidth]{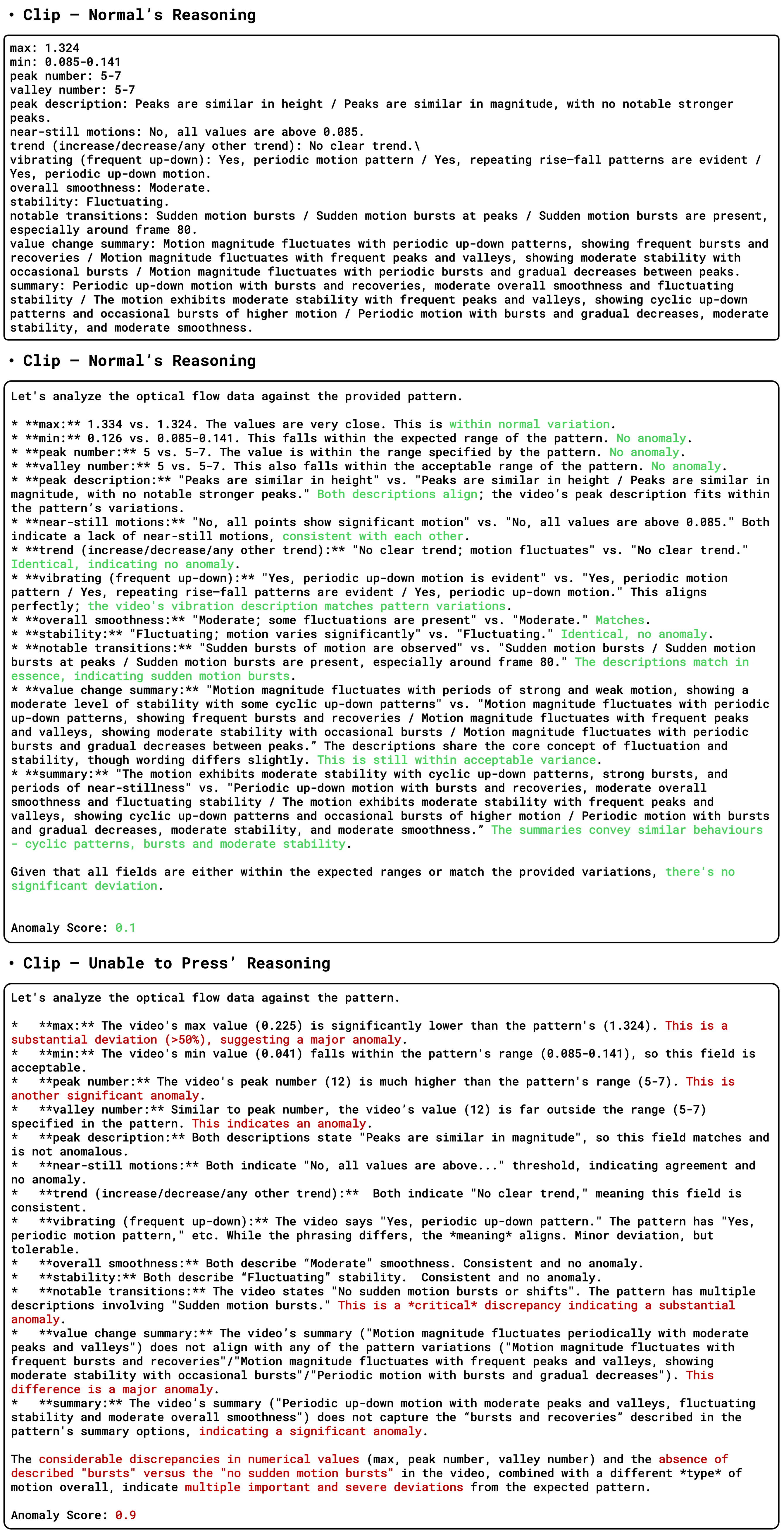}
    \caption{An example of the flow-magnitude-curve normal pattern and corresponding reasoning output.}
    \label{fig:oflow_pattern}
\end{figure*}

In Figure~\ref{fig:dynamic_pattern} and Figure~\ref{fig:oflow_pattern}, we show examples of trajectory frame description normal pattern and flow-mag curve normal pattern of rolling bearing and clip objects. We also show the reasoning results of both pattern with each a normal video and anomaly video frame descriptions. From the patterns and reasonings, we can see that our method can find the anomaly and give accurate anomaly scores to the videos.

\section{Comparison of Normal Representation Strategies}


Our framework extracts normal video descriptions to construct normal patterns. We compare against rule-based representations from proposed by Follow-the-Rules~\cite{yang2024follow}. To ensure fair evaluation, we apply both methods to the same trajectory frames and flow magnitude curves and with identical fusion procedures, and the result is in Table~\ref{tab:prompt_strategy}. The rule-based normal representation approach achieves an AUC of 0.527, compared to our pattern-based method's AUC of 0.681. This substantial performance gap demonstrates that extracted normal patterns capture motion characteristics more effectively than extracted normal rules. The superior performance stems from patterns' ability to model complex and subtle motion dynamics that rigid rule sets cannot adequately represent.

\begin{table}[ht]
\centering
\caption{AUROC on different normal representation Strategies}
\label{tab:prompt_strategy}
\begin{tabular}{lrr}
\toprule
Normal representation & Rules & Pattern (MUSTARD) \\
\midrule
AUC & 0.527 & 0.681 \\
\bottomrule
\end{tabular}
\end{table}

\section{Computational Efficiency}

\paragraph{\rev{Runtime Evaluation.}}
\rev{To evaluate inference efficiency, we sample 10 test videos from each of the
22 object categories in Phys-AD, resulting in 220 videos in total. We compare
MUSTARD with vanilla Qwen2.5-VL-7B inference using 8 frames (Qwen-8) and
40 frames (Qwen-40). As shown in Figure.~\ref{fig:efficiency}, Qwen-8,
Qwen-40, and MUSTARD require 12.0, 87.3, and 64.3 seconds per video,
respectively, while achieving AUROC scores of 0.500, 0.492, and 0.656.
These AUROC values are measured on the sampled 220-video subset and may
therefore differ slightly from the full-dataset results reported in the main
paper. We observe that the additional inference time required by MUSTARD corresponds to a measurable improvement in performance, whereas Qwen-40 incurs higher latency without a comparable gain. This demonstrates the effectiveness of MUSTARD's additional inference-time computation.}

\paragraph{\rev{Memory Footprint and Deployment.}}
\rev{As shown in Figure.~\ref{fig:efficiency} and
Table~\ref{tab:mustard_memory}, MUSTARD has a peak GPU memory footprint of
14.7~GB during pattern reasoning and fusion, compared with 18.6~GB for
Qwen-8 and 21.1~GB for Qwen-40. The MLLM/LLM stages are executed
sequentially, so the models do not need to reside in GPU memory simultaneously.
The optical-flow preprocessing is performed on the CPU at $640\times360$
resolution and processes consecutive frame pairs rather than loading the full
video into memory. For a representative 120-frame video with 4,000 tracked
points, the explicitly allocated frame, flow, and trajectory arrays require
approximately 15--30~MiB of RAM. This estimate excludes OpenCV's internal
TV-L1 optimization buffers, whose exact peak allocation depends on the OpenCV
implementation.}

\section{Alternative Backbone MLLMs}
\label{sec:stronger_mllm}



\rev{To demonstrate MUSTARD's generalizability across backbone MLLMs, we further evaluate it using InternVL3.5-8B as an alternative backbone. For each MUSTARD variant, the backbone MLLM describes all three modalities, while Gemma3-12B performs pattern extraction, reasoning, and multi-source reasoning fusion. This comparison isolates gains from stronger foundation models from those from MUSTARD's motion-aware representations and normal-reference reasoning.}

\rev{As shown in Table~\ref{tab:stronger_mllm}, applying MUSTARD yields substantial AUC gains for both MLLM backbones. Vanilla Qwen2.5-VL-7B and InternVL3.5-8B baselines achieve only 0.501 and 0.502 AUC, respectively; both are near chance level. Their corresponding MUSTARD variants, in contrast, reach 0.681 and 0.613. This substantial gain demonstrates that the proposed motion-aware representations and normal-reference reasoning are effective across different backbone MLLMs.}

\section{Examples of Prompts}

\begin{figure*}[!p]
    \centering
    \includegraphics[width=0.90\textwidth]{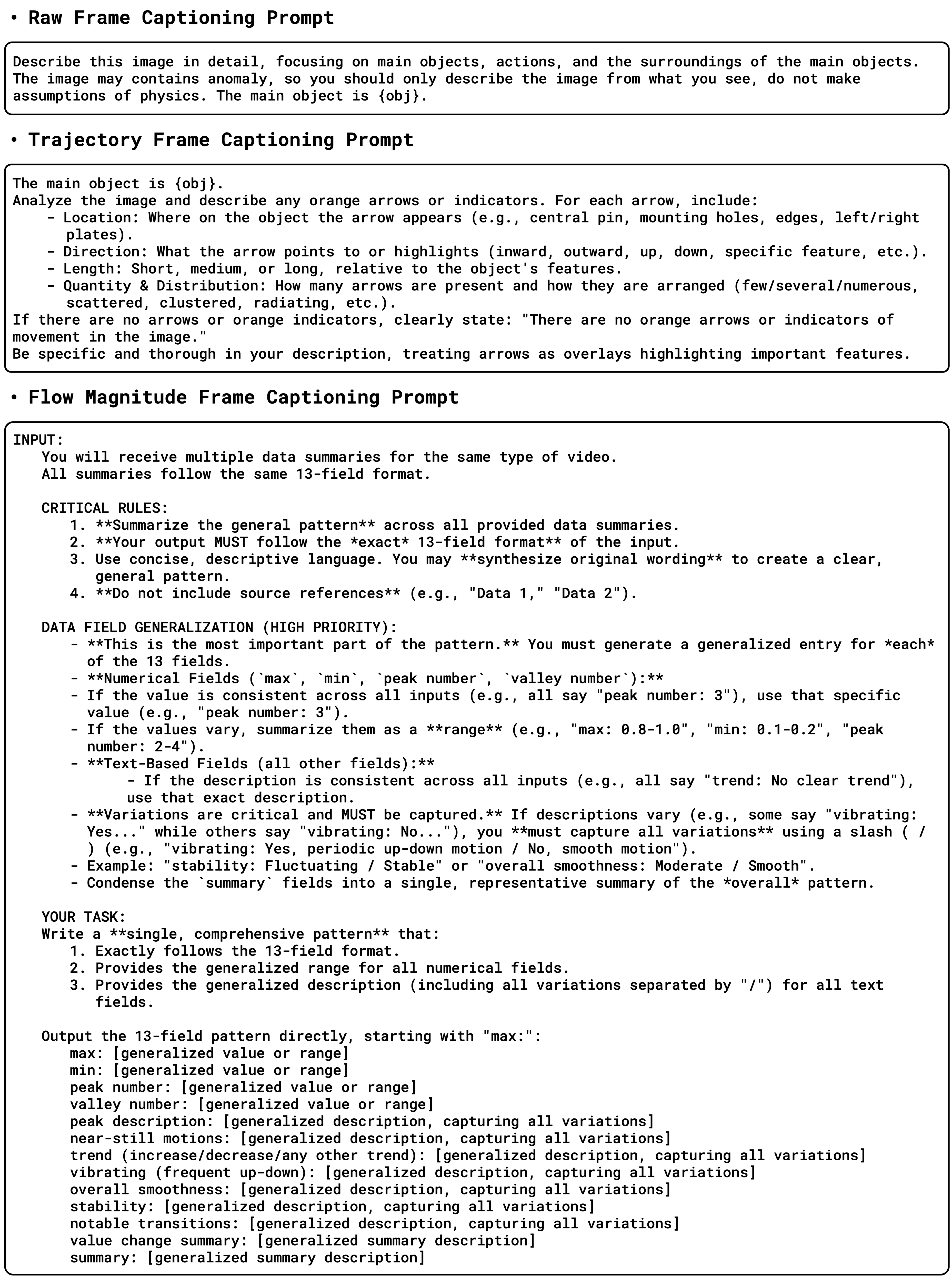}
    \caption{Prompts of each source fed to Qwen2.5-VL-7B to generate frame descriptions.}
    \label{fig:caption_prompt}
\end{figure*}

\begin{figure*}[!p]
    \centering
    \includegraphics[width=0.78\textwidth]{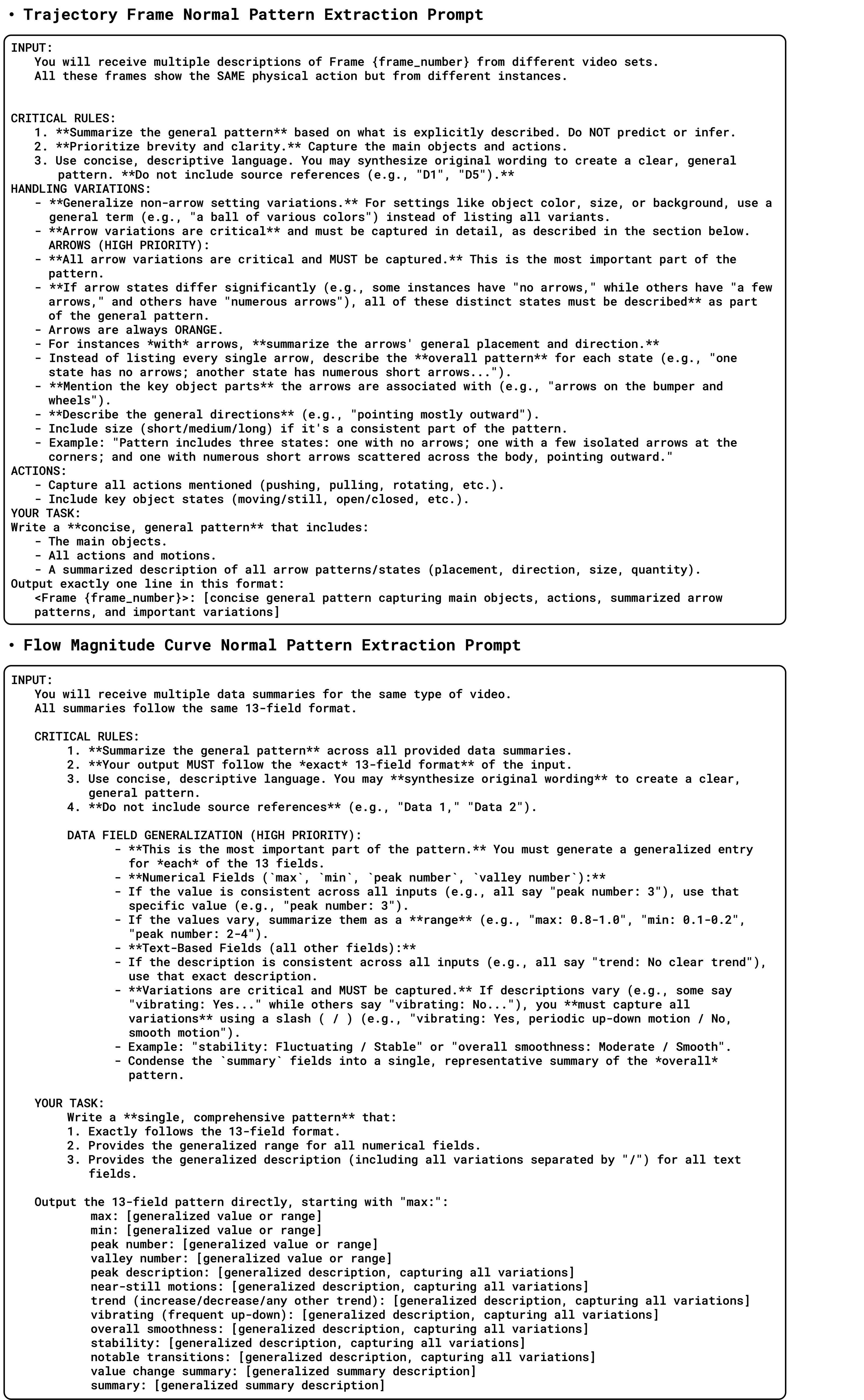}
    \caption{Prompts fed to Gemma3-12B to capture normal patterns of trajectory-frame descriptions and flow-magnitude curves.}
    \label{fig:pattern_extraction}
\end{figure*}

\begin{figure*}[!p]
    \centering
    \includegraphics[width=0.85\textwidth]{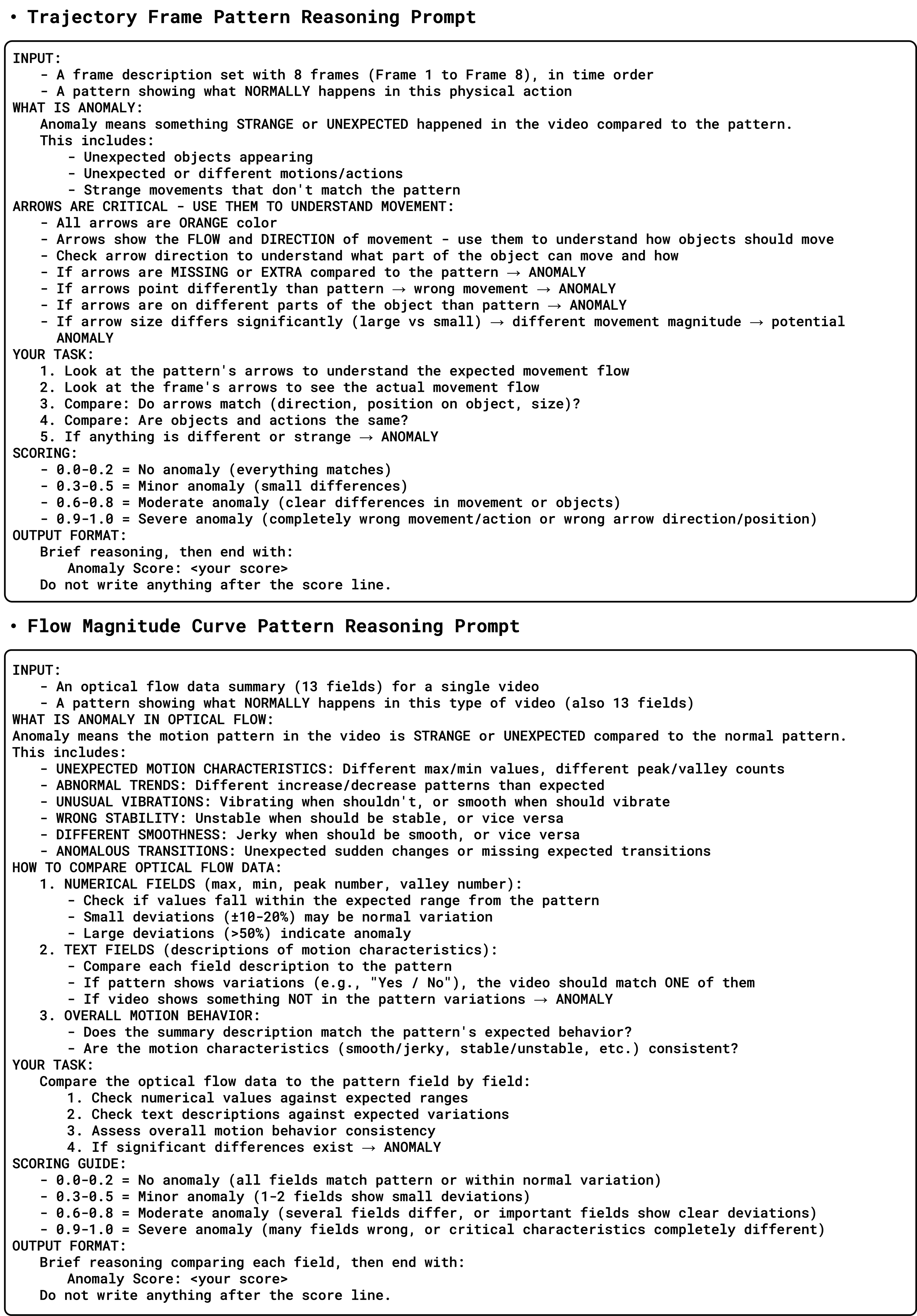}
    \caption{Prompts fed to Gemma3-12B to compare frame descriptions with the normal patterns.}
    \label{fig:pattern_reasoning}
\end{figure*}

\begin{figure*}[!p]
    \centering
    \includegraphics[width=0.85\textwidth]{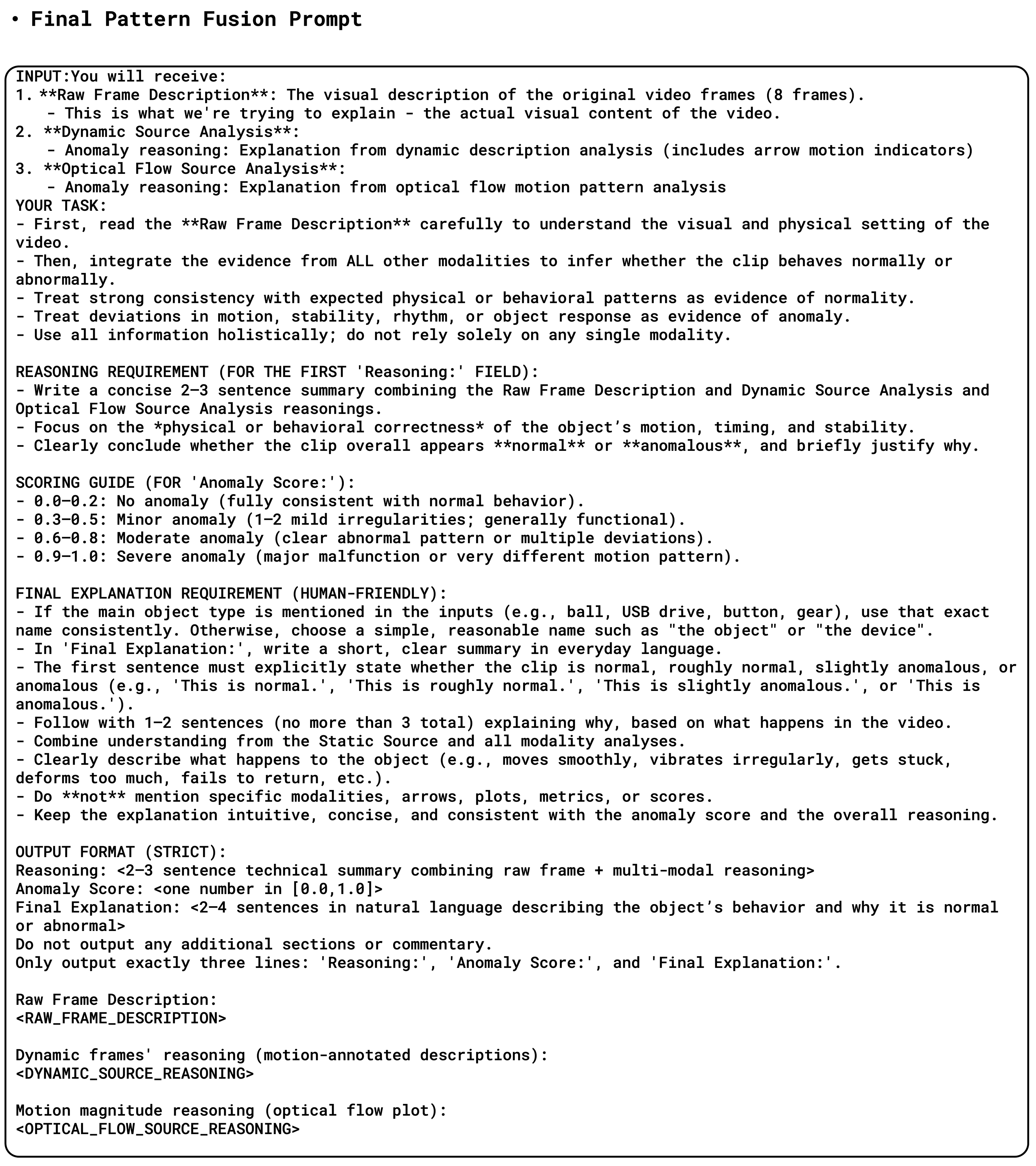}
    \caption{Prompt provided to Gemma3-12B for fusing reasoning from trajectory frames,
    the flow-magnitude curve, and raw-frame descriptions.}
    \label{fig:pattern_fusion}
\end{figure*}

\rev{We provide the complete prompts used throughout MUSTARD.
Figure.~\ref{fig:caption_prompt} presents the prompts for generating raw-frame,
trajectory-frame, and flow-magnitude descriptions.
Figure.~\ref{fig:pattern_extraction} presents the prompts for extracting normal
patterns from reference videos.
Figure.~\ref{fig:pattern_reasoning} presents the prompts for comparing test-video
descriptions against the extracted normal patterns and generating
modality-specific anomaly scores.
Finally, Figure.~\ref{fig:pattern_fusion} presents the prompt for fusing the
multi-source reasoning results into the final anomaly score and explanation.}


\vspace{0.5em}
\nolinenumbers

\noindent
\begin{minipage}{\columnwidth}
\centering

\captionsetup{
    font=small,
    skip=4pt
}
\captionof{table}{\rev{GPU memory footprint of the major MUSTARD inference stages.}}
\label{tab:mustard_memory}

\renewcommand{\arraystretch}{1.15}

\resizebox{0.8\columnwidth}{!}{%
\begin{tabular}{@{}lcc@{}}
\toprule
\textbf{Stage} & \textbf{Model} & \textbf{GB} \\
\midrule
Raw-frame Desc.    & Qwen2.5-VL-7B & 7.7 \\
Flow-mag Desc.     & Qwen2.5-VL-7B & 8.0 \\
Reasoning + Fusion & Gemma3-12B    & \textbf{14.7} \\
\midrule
\textbf{Peak} & -- & \textbf{14.7} \\
\bottomrule
\end{tabular}%
}

\end{minipage}

\vspace{0.45em}

\noindent
\begin{minipage}{\columnwidth}
\centering











\begin{tikzpicture}
\begin{axis}[
    width=0.90\linewidth,
    height=4.6cm,
    ybar,
    bar width=10pt,
    xmin=0.5,
    xmax=3.5,
    ymin=0,
    ymax=140,
    ytick={0,50,100},
    xtick={1,2,3},
    xticklabels={Qwen-8,Qwen-40,MUSTARD},
    ylabel={Runtime (s)},
    axis lines=box,
    tick label style={font=\scriptsize},
    label style={font=\scriptsize},
    nodes near coords,
    nodes near coords style={
        font=\scriptsize,
        anchor=south,
    }
]
\addplot coordinates {
    (1,12.0)
    (2,87.3)
    (3,64.3)
};
\end{axis}
\begin{axis}[
    width=0.90\linewidth,
    height=4.6cm,
    xmin=0.5,
    xmax=3.5,
    ymin=0.45,
    ymax=0.70,
    xtick=\empty,
    ylabel={AUROC},
    axis y line*=right,
    axis x line=none,
    tick label style={font=\scriptsize},
    label style={font=\scriptsize},
    legend style={
        at={(0.04,0.96)},
        anchor=north west,
        fill=white,
        font=\scriptsize,
        legend columns=1
    },
    legend cell align={left}
]
\addlegendimage{area legend, fill=blue!30, draw=blue!70!black}
\addlegendentry{Runtime}
\addplot[
    thick,
    mark=*,
    mark size=1.4pt
] coordinates {
    (1,0.500)
    (2,0.492)
    (3,0.656)
};
\addlegendentry{AUROC}
\end{axis}
\end{tikzpicture}

\vspace{-0.45em}
{\scriptsize (a) Runtime--accuracy trade-off.}

\vspace{0.2em}

\begin{tikzpicture}

\begin{axis}[
    width=0.72\linewidth,
    height=5cm,
    ybar,
    bar width=10pt,
    xmin=0.5,
    xmax=3.5,
    ymin=0,
    ymax=24,
    xtick={1,2,3},
    xticklabels={Qwen-8,Qwen-40,MUSTARD},
    ylabel={GPU Mem. (GB)},
    axis lines=box,
    tick label style={font=\scriptsize},
    label style={font=\scriptsize},
    nodes near coords,
    nodes near coords style={
        font=\scriptsize,
        anchor=south,
        yshift=1pt
    }
]

\addplot coordinates {
    (1,18.6)
    (2,21.1)
    (3,14.7)
};

\end{axis}

\end{tikzpicture}

\vspace{-0.45em}
{\scriptsize (b) Peak GPU memory.}

\vspace{0.15em}

\captionsetup{
    font=small,
    skip=2pt,
    justification=justified,
    singlelinecheck=false
}

\captionof{figure}{
Computational efficiency on the 220-video subset.
}
\label{fig:efficiency}


\end{minipage}

\linenumbers


\clearpage
\nolinenumbers

\begin{figure*}[!t]
\centering

\begin{minipage}{0.94\textwidth}
\centering

\small
\setlength{\tabcolsep}{7pt}
\renewcommand{\arraystretch}{1.05}

\captionof{table}{
\rev{Comparison with different backbone MLLMs on the Phys-AD dataset.
For MUSTARD variants, the description backbone is shared across raw-frame,
trajectory-frame, and flow-magnitude descriptions.
}}
\label{tab:stronger_mllm}

\vspace{0.25em}

\begin{tabular}{@{}lcccc@{}}
\toprule
\textbf{Method}
& \textbf{Description Backbone}
& \textbf{Pattern / Reasoning}
& \textbf{AUROC (\%)} \\
\midrule


Vanilla backbone MLLM
& Qwen2.5-VL-7B
& Direct
& 50.1 \\

MUSTARD
& Qwen2.5-VL-7B
& Gemma3-12B
& 68.1 \\

\midrule

Vanilla backbone MLLM
& InternVL3.5-8B
& Direct
& 50.2 \\


MUSTARD
& InternVL3.5-8B
& Gemma3-12B
& 61.3 \\


\bottomrule
\end{tabular}

\end{minipage}

\vspace{1.0em}

\includegraphics[width=0.78\linewidth]{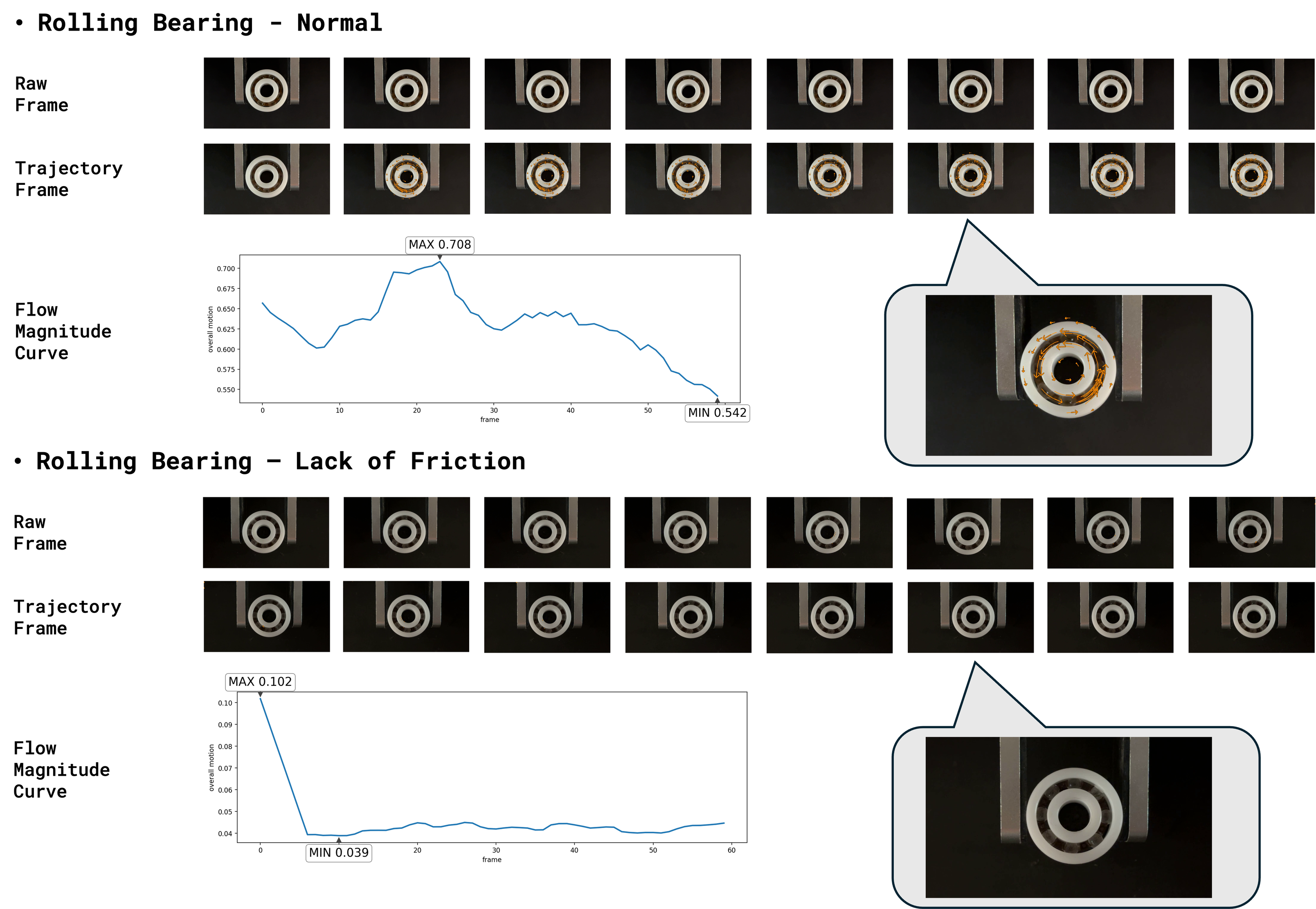}

\vspace{-0.2em}

\caption{An example of three different data sources for the Rolling Bearing object.}
\label{fig:rolling_bearing}

\end{figure*}

\linenumbers

\clearpage

\end{document}